\documentclass[lettersize,journal]{IEEEtran}
\usepackage{amsmath,amsfonts}
\usepackage{algorithmic}
\usepackage{algorithm}
\usepackage{array}
\usepackage[caption=false,font=normalsize,labelfont=sf,textfont=sf]{subfig}
\usepackage{textcomp}
\usepackage{stfloats}
\usepackage{url}
\usepackage{verbatim}
\usepackage{graphicx}
\usepackage{cite}
\usepackage{amssymb}
\usepackage{multirow}
\usepackage{booktabs}
\usepackage{colortbl}
\usepackage{xcolor}
\usepackage{array}
\usepackage{tabularray}
\usepackage{booktabs}    
\usepackage{pifont}      
\usepackage{multirow}    

\usepackage{marginnote}

\newcommand{\cmark}{\ding{51}}  
\newcommand{\xmark}{\ding{55}}  

\definecolor{mygray}{gray}{.9}
\definecolor{light-gray}{gray}{0.72}

\definecolor{citecolor}{RGB}{34,139,34}
\makeatletter\renewcommand\paragraph{\@startsection{paragraph}{4}{\z@}
	{.35em \@plus1ex \@minus.2ex}{-.5em}{\normalfont\normalsize\bfseries}}\makeatother
\usepackage{url}
\usepackage{enumerate}
\usepackage{enumitem}
\usepackage{xcolor}
\usepackage{booktabs}
\usepackage{hyperref}

\usepackage{xparse}   
\usepackage{hyperref} 

\hypersetup{
    colorlinks=true,
    linkcolor=blue,
    filecolor=blue,      
    urlcolor=blue,
    citecolor=blue,
}
\usepackage{comment}

\definecolor{newcolor}{rgb}{.8,.349,.1}

\definecolor{red}{rgb}{0.8,0,0}
\definecolor{blue}{rgb}{0,0,0.8}
\definecolor{green}{rgb}{0,0.4,0}

\newcommand{\change}[2]{}
\newcommand{\lchange}[2]{}

\begin{document}

\twocolumn
\pagenumbering{arabic}
\setcounter{page}{1}
\setcounter{table}{0}
\setcounter{figure}{0}
\setcounter{equation}{0}


\title{EndoSERV: A Vision-based Endoluminal Robot Navigation System}

\author{Junyang Wu, Fangfang Xie, Minghui Zhang, Hanxiao Zhang, Jiayuan Sun, Yun Gu,~\IEEEmembership{Member,~IEEE,} and Guang-Zhong Yang,~\IEEEmembership{Fellow,~IEEE}
\thanks{Junyang Wu, Minghui Zhang, Hanxiao Zhang, Yun Gu and Guang-Zhong Yang are with the Institute of Medical Robotics, Shanghai Jiao Tong University, Shanghai, China, 200240. (Email: yungu@ieee.org, gzyang@sjtu.edu.cn)}%
\thanks{Fangfang Xie and Jiayuan Sun are with the Shanghai Chest Hospital, Shanghai, China.}
}

\markboth{IEEE TRANSACTIONS ON ROBOTICS}%
{Wu \MakeLowercase{\textit{et al.}}: EndoSERV}


\maketitle

\begin{abstract}
Robot-assisted endoluminal procedures are increasingly used for early cancer intervention. However, the intricate, narrow and tortuous pathways within the luminal anatomy pose substantial difficulties for robot navigation. Vision-based navigation offers a promising solution, but existing localization approaches are error-prone due to tissue deformation, in vivo artifacts and a lack of distinctive landmarks for consistent localization. This paper presents a novel EndoSERV localization method to address these challenges. 
It includes two main parts, \textit{i.e.}, \textbf{SE}gment-to-structure and \textbf{R}eal-to-\textbf{V}irtual mapping, and hence the name.
For long-range and complex luminal structures, we divide them into smaller sub-segments and estimate the odometry independently. To cater for label insufficiency, an efficient transfer technique maps real image features to the virtual domain to use virtual pose ground truth. The training phases of EndoSERV include an offline pretraining to extract texture-agnostic features, and an online phase that adapts to real-world conditions. Extensive experiments based on both public and clinical datasets have been performed to demonstrate the effectiveness of the method even without any real pose labels.

\end{abstract}


\section{Introduction}
\IEEEPARstart{E}{ndoluminal} intervention is an effective tool for integrated diagnosis and treatment of early luminal cancers. It is increasingly used for digestive, pulmonary, urinary, and gynecologic tract diseases, offering a minimally invasive and one-stop-shop to alternative surgical procedures. The emergence of endoluminal robotics platforms further enhances the safety, consistency, maneuverability, and accuracy of endoscopic navigation, allowing small lesions to be targeted and dissected accurately.
For deep, complex, tortuous lumens, however, accurate navigation in a maze-like internal structure is a major challenge, which is further hampered by limited endoscopic field-of-view (FoV), in vivo artifacts such as blood, mucus, and motion blur, as well as constant tissue deformation and a lack of distinctive feature landmarks.

Existing endoluminal navigation techniques  can be broadly categorized as electromagnetic and shape-sensing based navigation, and pure vision-based navigation methods. Electromagnetic and shape-sensing based navigation depends on specialized external tracking devices.
Pure vision-based approaches, on the other hands, leverage real-time endoscopic images for guidance, offering a cost-effective and more flexible solution by eliminating the need for additional hardware that may complicate the existing surgical workflow. 

Existing vision-based methods can be classified into three main categories: \textbf{Structure-from-Motion (SfM)}, \textbf{Neural Implicit-based SLAM}, and \textbf{Real-Virtual Alignment} techniques.

\begin{figure}[!t]
\centering
\includegraphics[width=\linewidth]{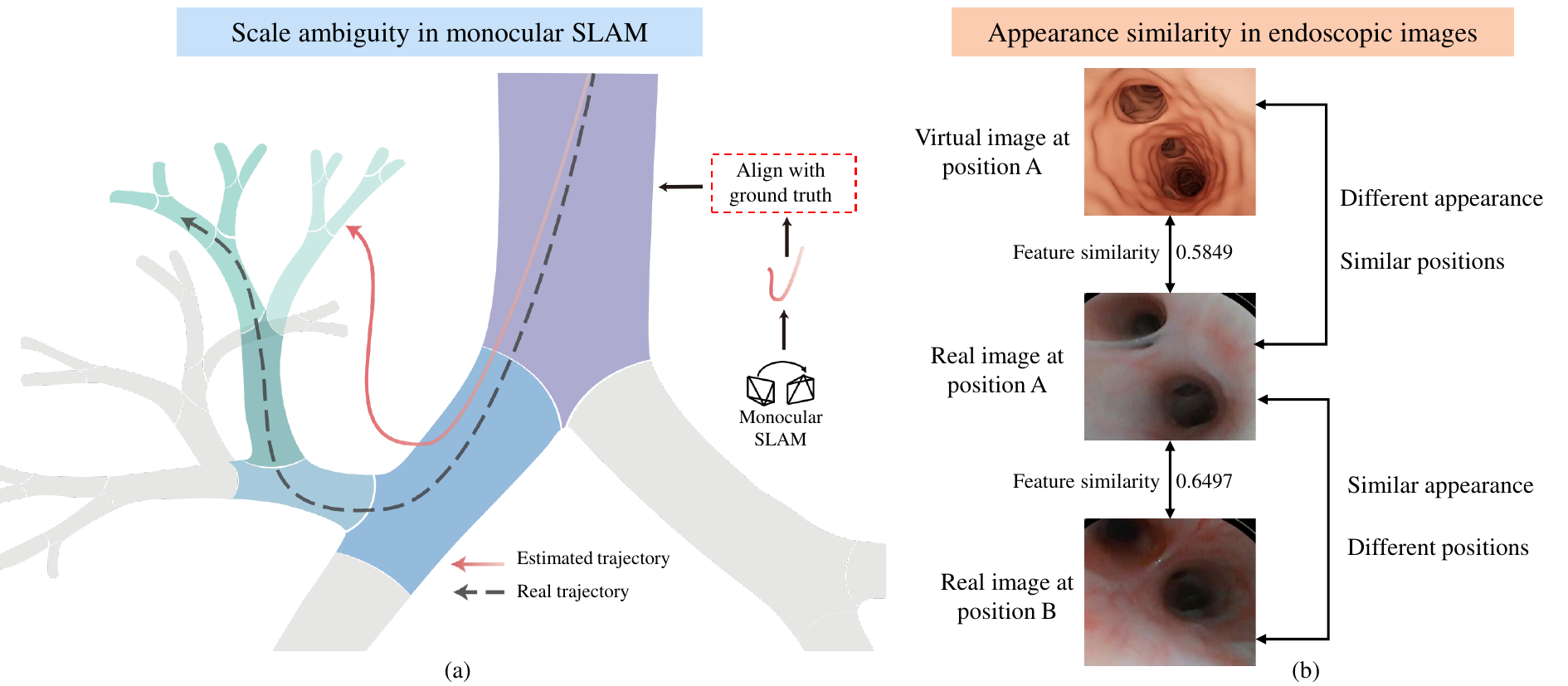}
\caption{ \textbf{(a). Scale ambiguity in monocular SLAM}: During the testing phase, the results of monocular SLAM require alignment with ground truth trajectories, which is impractical in clinical applications due to the lack of absolute scale information. \textbf{(b). Appearance similarity in endoscopic images}: Different bronchial branches often exhibit nearly identical geometric and topological structures, which can lead to ambiguous feature matching and incorrect associations with virtual frames. }
\label{fig:intro_chall} 
\end{figure}
 \textbf{SfM}: SfM methods use frame-to-frame correlations to estimate camera motion, leveraging appearance differences to optimize the pose estimation. By warping the previous frame to the next frame, the difference in appearance can be used as a constraint to optimize the camera pose. Previous studies, such as \cite{endoslam, liu2019dense, AF, tcl}, have explored SfM techniques in endoscopic scenarios. Cui et al. \cite{surgicaldino, endodac} introduced fine-tuning on top of Depth-Anything \cite{depthanything}, generating improved depth estimation results.
 However, for the monocular localization task, the scale information is limited. These methods needs the scale alignment with ground truth during testing phase, which is not accepted in the clinical scenarios. 
 
 \textbf{Neural Implicit-Based SLAM}: Neural implication-based SLAM is a promising approach that combines neural implicit networks with SLAM for camera pose estimation and mapping. Sucar et al. \cite{imap} and Johari et al. \cite{eslam} combined neural representation and SLAM to provide odometry estimation and 3D reconstruction. Shan et al. \cite{enerf} and Wang et al. \cite{endogslam} proposed dense SLAM systems using neural representations in endoscopic settings. However, methods based on neural implicit networks are time-consuming to render, making real-time navigation difficult. In addition, because these methods require high consistency among images within a scene, their performance is suboptimal in clinical scenarios when artifacts, tissue deformation and inter-reflections are abundant.

\begin{figure}[]
\centering
\includegraphics[width=\linewidth]{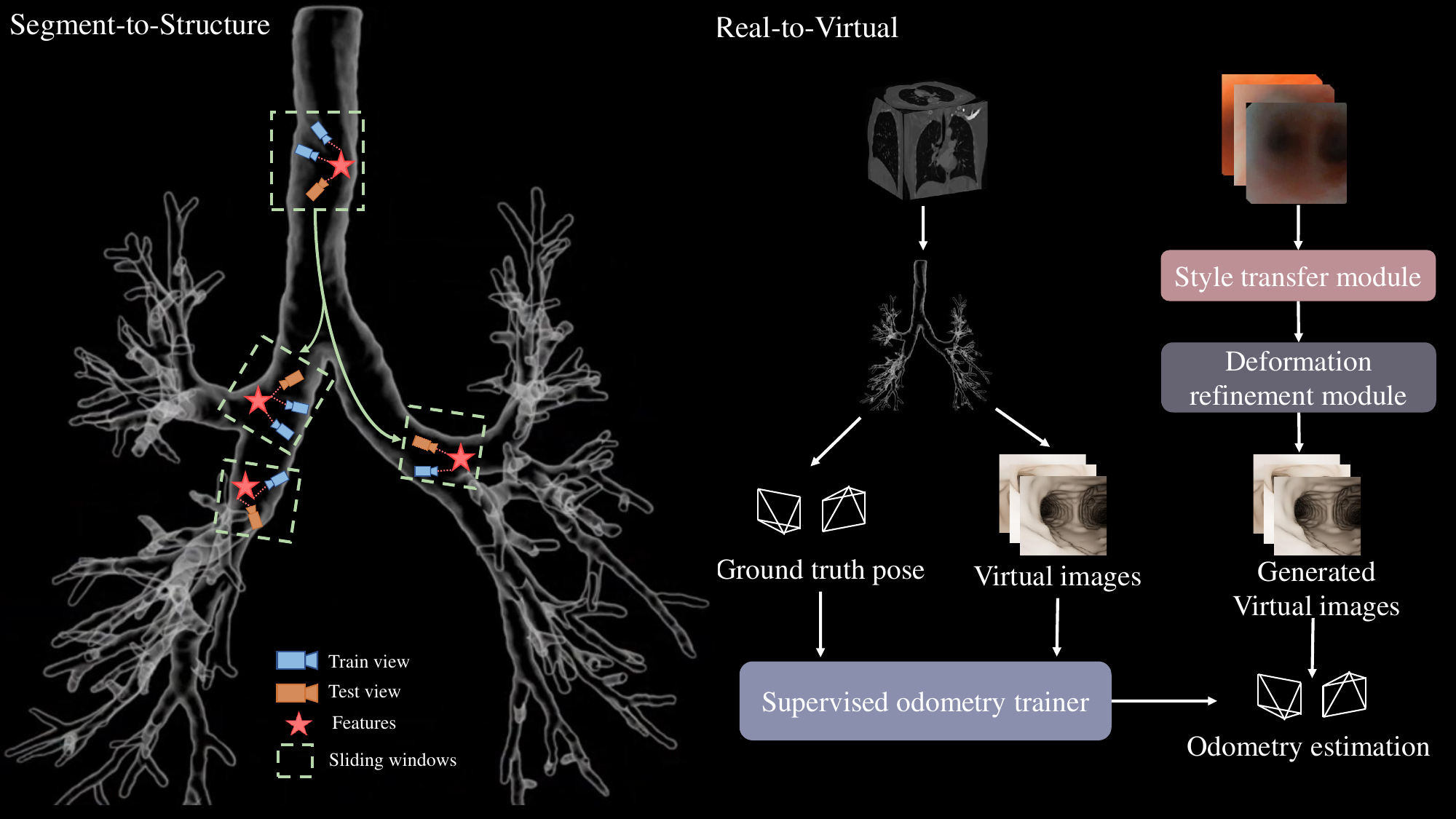}
\caption{ \textbf{Motivations of this work.} \textbf{Segment-to-Structure}: Due to the complex and long-term structure in the luminal path, a divide-and-conquer strategy is proposed for long-term pose estimation. \textbf{Real-to-Virtual}: Due to the lack of real pose labels in clinical scenarios, in this work, pre-operative CT data are used as the structure prior for intra-operative odometry estimation. }
\label{fig:intro} 
\end{figure}

\textbf{Real-Virtual Alignment}: 3D models reconstructed from pre-operative images such as CT and MRI can provide virtual images and corresponding poses, providing a strong prior for localization. Aligning real and virtual images enables the use of this pre-operative information for localization~\cite{luo2014discriminative,luo2012development}. 
Zhu et al. \cite{cyclegan_} and Park et al. \cite{cut} proposed image-to-image translation methods that can bridge the gap between real and virtual domains. Shen et al. \cite{shen2019context}, Gu et al. \cite{gu2022vision},  and Luo et al. \cite{luo2023monocular,luo2020new}, aligned the virtual and real domains at the feature level, measuring the similarity between real images and virtual images. The measurement can then be used to perform real/virtual registration. 

Despite promising results on simple datasets, existing endoscopic localization methods often exhibit challenges when applied to real-world clinical data. As shown in Fig. \ref{fig:intro_chall} (a), Structure-from-Motion (SfM) and neural implicit-based SLAM approaches estimate relative poses and lack access to absolute scale in monocular endoscopic settings, necessitating post-hoc alignment with ground truth trajectories. Such alignment procedures are infeasible in practical clinical deployments, where ground truth is unavailable. Moreover, the accumulation of errors over time in relative pose estimation leads to drift, further compromising navigation accuracy. While real-virtual alignment methods can estimate absolute poses by leveraging preoperative virtual images, their performance remains limited in clinical settings due to the inherent challenges of endoscopic images. Specifically, as shown in Fig. \ref{fig:intro_chall} (b), clinical endoscopic scenes often exhibit sparse textures, low contrast, and high visual similarity across different regions, resulting in localization ambiguities and disorientation during navigation.

To address these challenges, we propose a novel \textbf{EndoSERV} localization system, which includes two parts, \textit{i.e.}, \textbf{SE}egment to Structure and \textbf{R}eal to \textbf{V}irtual mapping. 
As illustrated in Fig. \ref{fig:intro}, the system adaptively divides the luminal pathway into manageable sub-segments, enabling independent analysis within each segment. Each segment is unique and do not confuse the localization system. In each sub-segment, a style transfer module and a deformation refinement module are used to adapt real images to the virtual domain, and an odometry trainer refines pose estimation using virtual labels to improve accuracy, while \textit{reducing the need for real labels}. Since the odometry network is trained in the virtual domain, which has the absolute scale information, the scale ambiguity can be solved.

The training pipeline comprises two key phases: an offline pretraining phase and an online training phase. During the offline phase, a texture-agnostic pose encoder is pretrained along with a foundational style transfer model. The encoder incorporates texture-diverse augmentations and texture-agnostic alignment strategies to ensure robustness across various textures. For the online phase, we optimize the localization module at the test time.
A feature retrieval step is used to identify a virtual buffer of frames relevant to the current real-world scenario. These frames, combined with real buffers, are used to efficiently fine-tune the localization system, ensuring real-time adaptability. To further enhance the performance under distortions and deformations in real-world scenarios, we propose an augmentation-then-recovery strategy for reconstructing virtual images from augmented real images. By aligning real and virtual data, our approach facilitates transfer to the virtual domain, enabling the scene coordinate network trained in the virtual domain to accurately estimate the camera pose in the real domain. In summary, EndoSERV enables endoscopic localization without requiring any real-world labels and demonstrates effectiveness compared to the current state-of-the-art on both public and clinical data.

\section{Related work}

Endoscopic navigation is a challenging task, primarily due to the absence of real-world labels for supervision. This section provides a detailed overview of existing work based on SfM, Neural-Implicit-based SLAM, and Real-Virtual Alignment and  related studies.

\subsection{SfM-based method}

SfM is one of the earliest and most widely applied self-supervised techniques for endoscopic navigation. It builds motion correlations between consecutive frames and optimizes camera poses using appearance differences as supervisory signals. EndoSLAM \cite{endoslam} is a seminal contribution that released large-scale datasets and formed the foundation for subsequent research. Liu et al. \cite{liu2019dense} further extended the application of SfM by training a convolutional neural network (CNN) with disparity supervision signals derived from conventional SfM algorithms, improving pose estimation accuracy. In addition, AF-SfMLearner \cite{AF} proposed an appearance module that mitigates brightness inconsistencies between adjacent frames, addressing one of the key challenges in SfM-based methods. TCL \cite{tcl} extended this idea by utilizing image triplets to increase the dataset diversity and reduce appearance inconsistencies in triplet images. With the rise of foundation models, SurgicalDINO \cite{surgicaldino} and EndoDAC \cite{endodac} fine-tuned depth estimation models like Depth Anything to improve depth prediction performance. However, although SfM-based methods can achieve real-time inference, they suffer from a lack of scale information and are unable to estimate absolute camera poses in real-world scenarios, limiting their applicability in clinical environments.

\subsection{Neural Implicit-based SLAM}
Neural implication-based SLAM is a promising alternative to traditional SfM-based methods. This approach utilizes implicit neural representations for camera pose estimation and scene mapping. iMap \cite{imap} was the first to integrate neural representations into the SLAM framework and demonstrated promising results. iNeRF \cite{inerf} employed a pretrained NeRF model to recover camera poses with greater accuracy. Subsequent works, such as eSLAM \cite{eslam} and COSLAM \cite{coslam} enhanced NeRF-based SLAM systems by implementing sparse parametric scene representations, thereby improving scalability and robustness. NICER-SLAM \cite{nicerslam} and DenseNeRF-SLAM \cite{densenerfslam} alleviated the dependence on RGBD inputs, enabling SLAM systems to operate using only RGB images. In the context of endoscopy, DenseSLAM systems based on neural representations, such as eNeRF \cite{enerf} and EndoGSLAM \cite{endogslam}, have demonstrated significant improvements in both mapping accuracy and pose estimation. Despite their promise, the computational cost of rendering images using NeRF models has prevented many of these systems from achieving real-time performance. Furthermore, the complex artifacts in endoscopic scenarios pose challenges to the application of NeRF-based methods in clinical settings.

\subsection{Real-Virtual Alignment}

In contrast to self-supervised methods, Real-Virtual Alignment bridges the gap between the intra-operative and pre-operative domains, taking advantage of pre-operative images such as CT and MRI as a prior for localization~\cite{luo2014discriminative,luo2012development}. 

Image-to-image translation is an efficient approach for Real-Virtual Alignment. It transforms the images from the source domain to the target domain.  pix2pix \cite{pix2pix} introduced a paired image translation method to transfer images from the source domain to the target domain. However, due to the scarcity of paired data, CycleGAN~\cite{cyclegan_} enabled unpaired image translation using cycle consistency. CUT \cite{cut} introduced a contrastive learning strategy that replaced the cycle-consistency constraint and achieved high-quality image translation. Similarly, MI2GAN \cite{MI2GAN} addressed the content distortion issue by proposing a disentangling strategy that preserves content information while translating textures. With the rapid development of diffusion-based models, image translation methods based on diffusion networks have become viable alternatives to GAN. The Denoising Diffusion Probabilistic Model (DDPM) \cite{diffusion} demonstrated its capability to progressively transform Gaussian noise into coherent signals. Subsequent investigations have further explored image translation tasks. UNIT-DDPM \cite{unit-ddpm} generated target domain images through a denoising Markov Chain Monte Carlo approach conditioned on the source images. EGSDE \cite{EGSDE} trained an assist function on both source and target domains and used it to assist energy-guided stochastic differential equations for realistic image generation. UNSB \cite{unsb} improved the simple Gaussian prior assumption by modeling a sequence of adversarial learning problems and achieved remarkable performance.

In endoscopic navigation, Mahmood et al. \cite{jhu} initiated the translation of real images into the virtual domain to enhance sparse medical datasets. Islam et al. \cite{augment} improved the approach by using cycle consistency loss and incorporating two discriminators to remove specular highlights in the virtual domain. CLTS-GAN \cite{cltsGAN} further refined techniques for controlling color, lighting, and texture in real endoscopic images, thereby accurately generating illumination information for real-world applications. 

Feature-level registration is another way to align real and virtual domains. It maps the images to a unified space and measures the similarity between virtual and real images. Shen et al. \cite{shen2019context} mapped all images to the depth domain, matching the depth maps between virtual and real images. Luo et al. \cite{luo2020new, luo2023monocular} proposed constrained evolutionary stochastic filtering, extracting stable features in the real domain, and combining features between virtual and real images. Zhu et al. \cite{zhu2024bronchoscopic} combined the NeRF SLAM and domain adaptation. It transferred the virtual images to the real domain and represented the 3D models as a neural radiance field using the generated real images and corresponding virtual poses. Subsequently, the camera pose was optimized to match the rendered images through NeRF and the real images.

\begin{figure*}[!t]
    \centering
    \includegraphics[width=\linewidth]{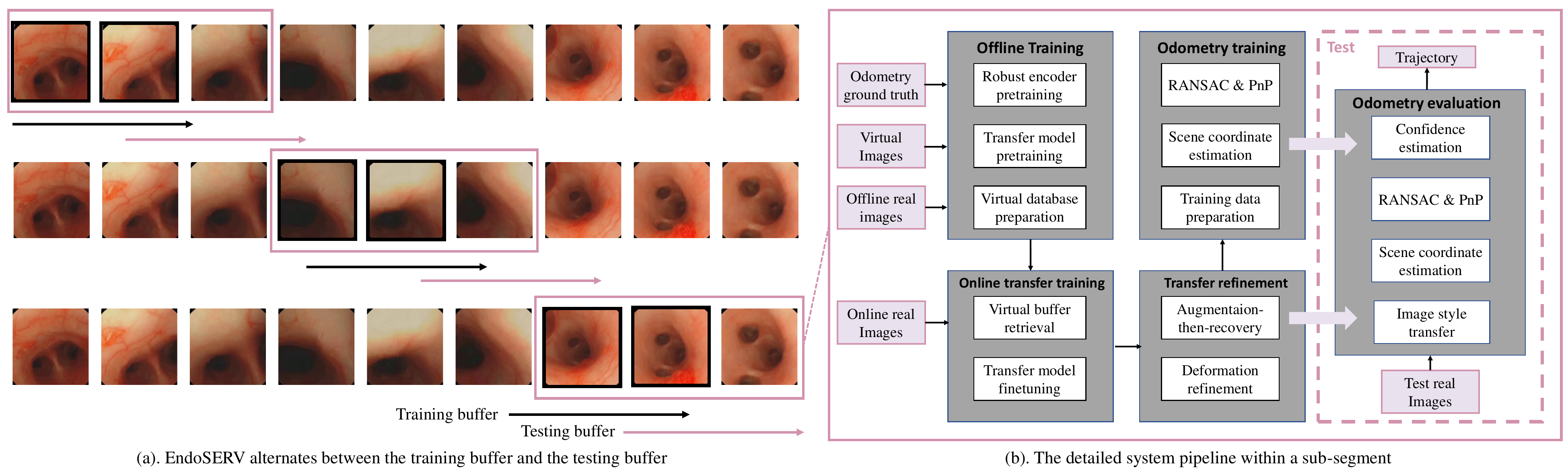}
    \caption{ \textbf{System overview of EndoSERV.} (a). A sliding windows strategy for long-term pose estimation. Black windows denote the training images, while pink windows represent the testing images. The sliding windows move along the temporal axis, enabling the system to alternate between training buffers and testing buffers. (b). The detail pipeline within a sub-segment, consisting of offline training, online training, and testing phase.
    }
    \label{fig:sliding window} 
    \end{figure*}

\section{System overview}
\subsection{Sliding Window Buffering}\label{subsec::sliding_window}

The endoluminal pathway is a highly complex and maze-like structure characterized by anatomically similar features across various regions. This inherent similarity among anatomical landmarks poses significant challenges for navigation algorithms, which can easily become misled, causing the system to lose direction during the navigation process.

To mitigate these challenges, we introduce a divide-and-conquer approach that utilizes sliding windows to localize within the endoluminal pathway. By partitioning the entire pathway into smaller sub-segments, each can be processed independently for greater precision and efficiency. As shown in Fig. \ref{fig:sliding window} (a), the system alternates between training and testing phases. We maintain a confidence buffer to adaptively switch between these two phases. In the training phase, a training buffer containing a few local samples is used to rapidly fine-tune the system for the current sub-segment. The system then transitions to the testing phase, performing pose estimation. This continues until the confidence buffer detects a significant drop in prediction confidence. Such a decrease indicates that the module requires optimization to adapt to the new data. This triggers the system to automatically revert to the training phase, constructing a new training buffer for the next sub-segment. This adaptive approach ensures the system maintains both accuracy and reliability throughout the navigation process, optimizing the balance between real-time efficiency and model performance.

\subsection{Subsegment system pipeline}\label{subsec::subseg}

Fig.\ref{fig:sliding window} (b) illustrates the detailed system pipeline, which is divided into offline training, online training, and testing phases. Specifically, offline training is conducted during the pre-operative stage, while the online training and testing phases are conducted in the intra-operative phase.

\subsubsection{Offline Training} 
The pipeline begins with \textbf{Robust Encoder Pretraining} (see Section \ref{subsec::offline_training_feature_extraction}), which trains a texture-agnostic pose encoder capable of extracting robust features. This is followed by \textbf{Transfer Model Pretraining} (see Section \ref{subsec::offline_training_feature_extraction}), where a style transfer model is trained using unpaired datasets to effectively handle domain gap. During the \textbf{Virtual Database Preparation} step, a comprehensive virtual database is constructed by collecting all virtual images. Features from these images are extracted using R2Former \cite{zhu2023r2former} and stored in a virtual feature database to support the following steps.

\subsubsection{Online Training} 
In the online training phase, the system adapts the offline-trained models to real-world conditions. First, \textbf{Virtual Buffer Retrieval} (see Section \ref{subsec:virtual_buffer_retrieval}) narrows the virtual database by retrieving regions that are more closely aligned with the distribution of the real images. This step ensures that the virtual images in the virtual buffer are contextually similar to real-world images encountered during navigation.

The virtual buffer is then used to \textbf{fine-tune the transfer model} (see Section \ref{subsec:transfer_model_finetune}), improving its robustness in handling real-world scenarios. Subsequently, the system applies \textbf{Deformation Refinement} (see Section \ref{subsec:deformation_refinement}), which employs an augmentation-then-recovery strategy. This strategy enhances the ability of the system to overcome distortions and deformations by augmenting the training data and recovering original virtual data. Once real and virtual images are aligned using the refined transfer model, the system trains a \textbf{Scene Coordinate Estimation} (see Section \ref{subsec:sc_estimation}) model that accurately predicts camera poses based on virtual image features.

\subsubsection{Testing Phase} 
During testing, the trained transfer model and scene coordinate estimation network are deployed on real images to estimate the camera pose. The system incorporates a \textbf{confidence estimation step} to evaluate the reliability of each predicted pose. If the confidence is high, testing continues seamlessly. In cases where the confidence drops significantly, the system reverts to the training stage to update the models.

\section{Training pipeline}
\subsection{Offline Training: Extracting Robust Features}\label{subsec::offline_training_feature_extraction}

\begin{figure*}[!t]
\centering
\includegraphics[width=\linewidth]{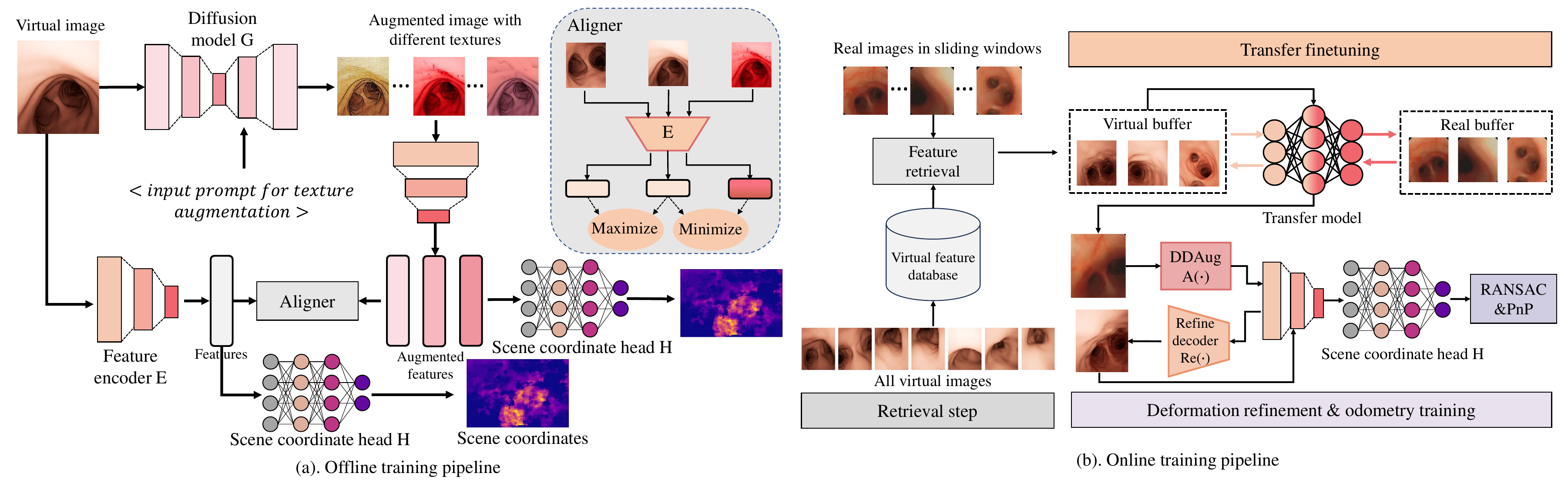}
\caption{ \textbf{Training pipeline overview.} (a). Offline training pipeline. Virtual images are augmented using the pretrained diffusion model, generating texture-diverse augmented images. A novel aligner is designed to constrain the feature encoder to extract features into a unified feature space. Scene coordinate head is designed to generate the scene coordinate map. (b). All virtual images are first compressed to the virtual buffer during a retrieval process, which is used to fine-tune the transfer model quickly with the real buffer. An augmentation-then-recovery strategy is proposed to refine the distortion and deformation issue. 
After aligning everything to the virtual domain, a scene coordinate head is trained to estimate the camera pose.
}
\label{fig:offline} 
\end{figure*}

The offline training pipeline, shown in Fig.\ref{fig:offline} (a), is designed to extract texture-agnostic features for robust pose estimation. To achieve this, we propose a pre-training framework that incorporates two key components: diverse texture generation and feature alignment.

A key insight for the generation of diverse textures is to generate images with stable structures and diverse textures. To accomplish this, we leverage a pre-trained diffusion model \cite{islam2024diffusemix} to generate augmented images, which preserve the structure but introduce texture variations.

To generate semantically meaningful prompts, we first leverage the advanced capabilities of the current large-language model, GPT-4o \cite{gpt4-o}, to generate prompts. We pose the following question:``Please describe the color and texture characteristics of endoscopic images.''  In response, GPT-4o generates set prompt $P = \{ p_{1}, p_{2}, \dots, p_{k} \}$. For the input image $I_{i}$, the generation step can be formulated as $I_{ij}^{g} = G(I_{i}, p_{j})$, where $G$ is the pretrained diffusion model and $p_{j}$ is a random prompt in the prompt sets. For the input virtual image $I_{i}$ and its augmentation $\{ I_{ij}^{g} \}_{j=0,\dots,k}$, the feature encoder extracts the corresponding features $F_{i} \in \mathbb{R}^{H \times W \times C}$ and $\{F_{j} \in \mathbb{R}^{H \times W \times C}\}_{j=0,\dots,k}$.

To align these features into a unified feature space, we employ two loss functions: similarity loss and contrastive loss. The similarity loss measures the cosine similarity between the feature representation of the virtual image and that of the augmented images, with the aim of minimizing this distance:
\begin{equation}
    L_{\textrm{sim}} = \sum_{j,h,w}(1 - \cos \frac{F_{i,h,w} \cdot F_{j,h,w}}{||F_{i,h,w}|| \cdot ||F_{j,h,w}||}) / (k \times H \times W)
\end{equation}
where $(h,w)$ is the corresponding feature location.

The contrastive triplet loss, on the other hand, encourages the network to minimize the distance between pairs of images with the same content but different textures, while maximizing the distance between pairs with identical textures but differing content. We sample a negative pair from another virtual image $I_{n}$, with the feature $F_{n}$ located at a different position. The contrastive triplet loss is then formulated as:

\begin{equation}
    L_{\textrm{triplet}} = \max\{ d(F_{i}, F_{j})  - d(F_{i}, F_{n}) + \tau, 0  \}
\end{equation}
where d($\cdot$) is the cosine distance, and $\tau$ is the margin value.

After feature extraction, a scene coordinate head $H$ is employed to estimate the scene coordinate of the image $I_{i}$.

Overall, during the offline training phase, the objective function is:

\begin{eqnarray}
\begin{aligned}
    L_{\textrm{offline}} =& L_{\textrm{sim}} + L_{\textrm{triplet}} + \\
    &L_{\textrm{proj}}(H, E, I_{i}, p^{*}) +  \frac{\sum_{j}L_{\textrm{proj}}(H, E, I_{j}, p^{*})}{k}
\end{aligned}
\end{eqnarray}
where $L_{\textrm{proj}}$ is the reprojection loss, which will be discussed in the following section, and $p^{*}$ is the virtual pose ground truth.  $E$ is the feature encoder.

Additionally, we pretrain a style transfer model using unpaired data to handle domain gap between real and virtual domains.

\subsection{Online Training: Adaptation to Real-World Scenarios}

While offline pretraining bridges the virtual and real domains, it may encounter limitations in handling real-world image deformations and artifacts. To address these challenges, we introduce a novel online training module designed to dynamically fine-tune the network and achieve reliable odometry estimation. This module focuses on rapid adaptation to environmental variations and refinement of distortion and deformation.

\subsubsection{Virtual Buffer Retrieval}
\label{subsec:virtual_buffer_retrieval}

The virtual database used for offline training is inherently large and time-consuming. Instead of relying on this exhaustive  database during online training, we introduce a retrieval algorithm to facilitate more targeted and efficient training.  
By matching the real images within the buffer with relevant data from the virtual database, we significantly reduce the size of the database that needs to be considered during training.

Specifically, for a training buffer $I_{t} = \{I_{1}, I_{2}, \dots, I_{T}\}$ and virtual database $\{I_{k}^{v}\}_{k=0,1,\dots,K}$, we perform a feature retrieval operation using R2former \cite{zhu2023r2former} as the feature extractor against the entire virtual database. This retrieval yields a list of indices $Idx = \{ idx_{1}, idx_{2}, \dots, idx_{T}\}$, representing that for the given real training image $I_{i}$, the most similar image from the virtual database is  $I_{idx_{i}}^{v}$.

After that, we define a Retrieval Hit Score $S_{k}$ for each virtual image $I_{k}^{v}$, representing the retrieval hit number of each virtual image:
\begin{equation}
    S_{k} = \sum_{i=1}^{T}1[k = Idx_{i}]
\end{equation}
where $1[\cdot]$ is the indicator function that is 1 if the virtual image $I_{k}^{v}$ appears in the retrieval indices $Idx$, and 0 otherwise. 

Furthermore, we define a contiguous range of $R$ virtual images in the virtual database that maximizes the total Retrieval Hit Score from the training buffer. 
For a contiguous subrange of $R$ virtual images, the range score can be defined as:
\begin{equation}
    S_{\textrm{range}}(k,k+R) = \sum_{j=k}^{k+R}S_{j}
\end{equation}

We then select the subrange of $R$ consecutive virtual images that maximizes the total retrieval hits:
\begin{equation}
    (k^{*}, k^{*} + R) = \arg \max_{k} S_{\textrm{range}}(k,k+R)
\end{equation}
where $(k^{*}, k^{*} + R)$ represents the optimal contiguous subrange of $R$ virtual images in the virtual database, which corresponds to the virtual buffer. This subrange is selected since it contains the highest concentration of retrieval hits from the training buffer images, ensuring that the virtual buffer is the most representative subset of the virtual database with respect to the current training buffer.

\subsubsection{Transfer Model Fine-tuning}
\label{subsec:transfer_model_finetune}

Upon selecting the virtual buffer, the next step involves fine-tuning the transfer model with both the real and virtual buffers.
During the offline training phase, the transfer model has already established a correspondence between the real and virtual domains.
At this stage, only the specific characteristics of the current environment need to be modeled, enabling rapid and effective fine-tuning.

\subsubsection{Deformation Refinement}
\label{subsec:deformation_refinement}

\begin{figure}[]
\centering
\includegraphics[width=\linewidth]{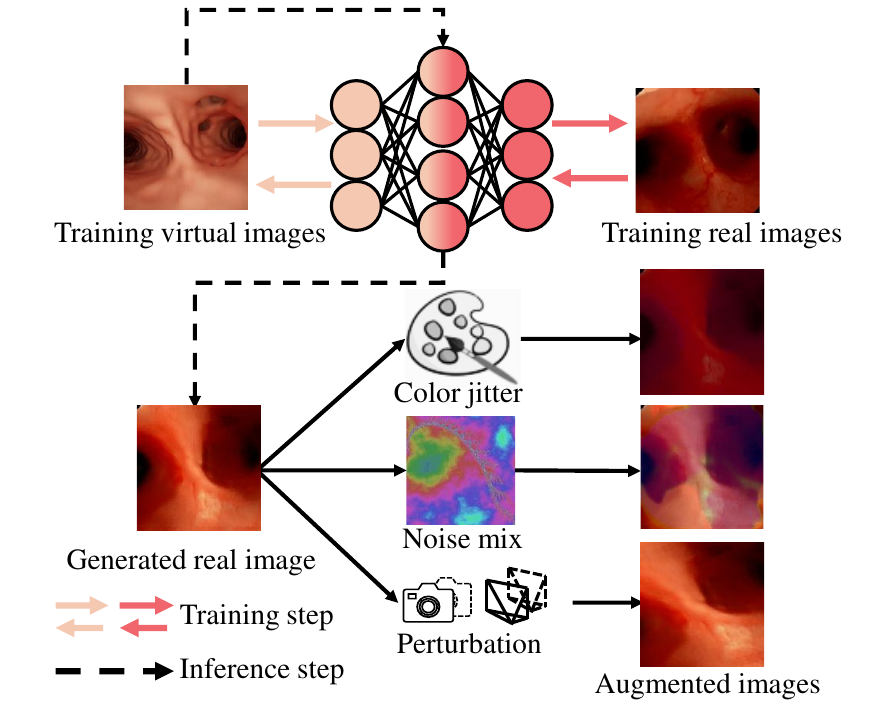}
\caption{ \textbf{DDAug framework.} The real image is generated from the virtual image using the pretrained transfer model. Three augmentations are applied: Color jitter, mixup with the noisy image, and camera parameter perturbation.
}
\label{fig:dda} 
\end{figure}

Although the transfer model bridges the virtual and real domains, it is limited by the absence of paired data. The unpaired nature of training results in insufficient precision, particularly for fine-grained tasks like pose estimation. Additionally, real images may contain artifacts, distortions, and deformations, further complicating accurate localization.

To address these issues, we introduce a novel augmentation-then-recovery strategy to refine the distortion and deformation issues. 
Specifically, we first design diverse augmentations to simulate the realistic image distortions and deformations, and recover the original images using a paired training pipeline.

As shown in Fig.\ref{fig:offline} (b), we begin by applying reverse transfer to generate real images from the virtual domain:$I_{r} = G(I_{v})$, where $I_{v}$ is the virtual images from the virtual buffer, $G$ is the reverse generator of the pretrained transfer model.

Though transfer models can mitigate the domain gap between real and virtual endoscopic data, real endoscopic images encountered in clinical scenarios often exhibit substantial distortions and deformations. For instance, artifacts such as bleeding, bubbles, and mucus can introduce significant noise, while inherent camera distortions may lead to pronounced scene deformations. To more realistically simulate the diverse conditions prevalent in clinical scenarios and further bridge this real-to-virtual domain gap, we propose a novel data augmentation method, termed DDAug. This approach incorporates three meticulously designed augmentation strategies, specifically engineered to emulate common occurrences in real endoscopic scenes, thereby enhancing the robustness and clinical applicability of the proposed system. Specifically, as shown in Fig.\ref{fig:dda}, three augmentations are applied: color jitter, noise mixup, and camera parameters perturbation.

\textbf{Color jitter}: We use traditional color jitter in this work, which introduces random changes in the image's brightness, contrast, and saturation to simulate variations in lighting.

\textbf{Noise mixup}: To simulate the artifacts in the endoscopy scenarios, we apply mixup between generated real images and fractal images \cite{islam2024diffusemix}, which are used for inducing structural variations in the hybrid images. A randomly selected fractal image $I_{F}$ is blended with the generated real images $I_{r}$ with a random blending factor $\lambda$: $I_{aug} = \lambda I_{r} + (1 - \lambda) I_{F}$.

\textbf{Camera parameters perturbation}: In this work, we apply camera parameters perturbation to simulate the deformation occurring in the real scenarios. Unlike traditional approaches that only adjust camera poses \cite{pda}, this work perturbs both the camera pose and intrinsic parameters, re-projecting all pixels to generate a novel synthetic image. This dual-parameter adjustment expands the diversity of training data.
Specifically, given a generated real image $I_{r}$, depth map from the virtual ground truth $D$, and intrinsic matrix $K$, we generate a new image $I_{p}$ by applying perturbations to both the rotation matrix and intrinsic parameters. Let $T_{p}$ represent the relative rotation matrix between the original and perturbed poses, and let $K_{p}$ denote the perturbed intrinsic matrix.
For any point $p = [u, v ]^{T}$ in $I_{r}$, we map it to a corresponding point $p^{'}$ via the following homography:
\begin{equation}
    p_{h}^{'} = K_{p}T_{p}K_{p}^{-1}zp_{h}
\end{equation} 
where z is the depth at position $p$, $p_{h}$ and $p_{h}^{'}$ represent the homogeneous coordinates of points $p$ and $p^{'}$. For the $T_{p}$ perturbation, we randomly sample pitch, yaw, and roll angles within the range of [-0.1,0.1] radians. Similarly, for the $K_{p}$ perturbation, we vary the intrinsic camera parameters $f_{x}$, $f_{y}$, $c_{x}$, and $c_{y}$ by up to 10\% of their respective values.

After the augmentations, we recover the original virtual images using a paired training pipeline.
For each augmented real image $I_{aug}$, the corresponding virtual image $I_{v}$ can be considered as the paired ground truth. Based on it, we employ a reconstruction decoder to map augmented image $I_{aug}$ to the virtual image $I_{v}$. RMSE is used as the loss function: 
\begin{equation}
    L_{\textrm{recon}} = || I_{v} - Re(E(A(G(I_{v}))))  ||_{2}
\end{equation}
where $G(\cdot)$ is the reverse generator of the pretrained transfer model, $A(\cdot)$ is the augmentation function, $E(\cdot)$ is the pretrained encoder, and $Re(\cdot)$ is the reconstruction decoder.

\subsubsection{Scene Coordinate Head Training}
\label{subsec:sc_estimation}

After aligning all data into the virtual domain, we apply a scene coordinate head and use the PnP algorithm to obtain the camera pose.

For the RGB image, we denote the scene coordinate $y_{i} \in Y$ as associated with pixel $x_{i}$, where the 2D-3D correspondences can be represented as: $C^{RGB} = \{ (x_{i},y_{i}) | y_{i}\in Y\}$. The relation between pixel coordinates and scene coordinates can be represented by $x_{i} = Kh^{-1}y_{i}$, where $K$ denotes the camera intrinsic matrix, and $h$ is the ground truth camera pose.

For estimating the scene coordinate, we employ an encoder-decoder neural network $f() = H(E(\cdot))$, where the encoder $E(\cdot)$ is fixed, and the scene coordinate head $H(\cdot)$ can be optimized. Following previous work which patchifies the image \cite{ace}, we estimate $y_{i} = f(p_{i};w)$, where $p_{i} = P(x_{i}, I)$ represents an image patch extracted around pixel position $x_{i}$ from the input image, and $w$ are the learnable parameters. The function $f$ maps the image patch to a 3D coordinate: $f: \mathbb{R}^{C_{1}\times H_{P} \times W_{P}} \rightarrow \mathbb{R}^{3}$, where $C_{1} = 1, H_{P} = W_{P} = 81$.
The objective function can be defined as: 

\begin{equation}
    L_{\textrm{proj}} = \sum_{i} ||p_{i} - Kh^{-1}f(p_{i};w)||_{2}
\end{equation}

Notably, pose ground truth $h$ is available exclusively for the virtual domain. The aforementioned scene coordinate head training is conducted entirely in the virtual domain. In summary, EndoSERV enables the training of a scene coordinate network and subsequent endoscopic localization without requiring any real-world labels.

\subsection{Confidence-Aware Pose Estimation}

During the inference phase, the aim is to recover camera pose $h$ from scene coordinate $Y$. 
This is achieved by employing a traditional Perspective-n-Point (PnP) minimal solver within a RANSAC loop. 
The pose hypothesis $h_{i}$ that maximizes consensus among the scene coordinates is selected as the final estimate:
\begin{equation}
    \tilde{h} = \arg \max_{h_{j}} s(h_{j}, Y)
\end{equation}

Here, $s(\cdot)$ is the scoring function, which is based on inlier counting:

\begin{equation}
    s(h, Y) = \sum_{y_{i} \in Y} 1[r(y_{i}, h)  \textless \tau ]
\end{equation}
where $r(\cdot)$ is the residual measurement function, $1[\cdot]$ is the indicator function.

Apart from estimating the camera pose, another key issue is to assess the reliability of pose estimation, particularly in the context of real-world applications such as medical robotic systems. For instance, if the surgeon remains stationary, the network can maintain high-confidence testing without needing to switch to the training phase. In contrast, rapid movement or larger scene variations may require a switch to the training phase for the system to learn from the new scene context. Thus, the ability to gauge the current estimate’s confidence is crucial for adaptive decision-making.

Building on this, we propose an extended use of inlier counts, not only for hypothesis selection but also for confidence estimation. 
In our experiments, we observe that inlier counts exhibit distinct patterns depending on the input data: high inlier counts are typically found in cases seen during training, while out-of-distribution (OOD) cases result in notably lower inlier counts. This observation motivates us to extend the role of inlier counting, not only for hypothesis selection but also for confidence estimation, allowing the system to determine when it can test with confidence and when it should transition to the training phase.

Specifically, during inference, we maintain a confidence buffer of fixed length (50 frames) that stores the inlier counts from recent frames. 
For each incoming frame, we compare its inlier count against the statistical thresholds of the buffer. If the inlier count of the current frame exceeds $\mu - 2 \sigma$($\mu$ and $\sigma$ are  the mean and standard deviation of inlier counts within the buffer), the frame is considered confident and is added to the buffer. Conversely, if the inlier count falls below this threshold, the frame is deemed uncertain and excluded from the buffer. If more than 20 frames are classified as uncertain, the system transitions from the testing phase to the online training phase.

\begin{table}
\centering
\caption{Pose estimation results of C3VD and clinical experiments}
\label{tab:synthetic}
\begin{tblr}{
  cells = {c},
  cell{1}{1} = {c=2,r=2}{},
  cell{3}{1} = {r=3}{},
  cell{6}{1} = {r=2}{},
  cell{8}{1} = {r=4}{},
  vline{2} = {2-11}{},
  vline{3-4} = {1-11}{},
  hline{1,3,6,8,12} = {-}{},
  hline{2} = {3-4}{},
  hline{11} = {2-4}{},
}
Method        &               & C3VD dataset  & Clinical data    \\
              &               & ATE($mm$)       & ATE($mm$)          \\
SfM           & AF-SfMLearner~\cite{AF} & 4.69$\pm$2.18 & 12.49$\pm$5.06   \\
              & EndoDAC~\cite{endodac}       & 3.89$\pm$2.50 & 11.10$\pm$5.63  \\
              & LightMono~\cite{luo2023monocular}     & 4.81$\pm$3.40 & 12.26$\pm$5.97   \\
NeRF-SLAM     & EndoGSLAM~\cite{endogslam}     & 4.35$\pm$1.90 & 12.04$\pm$5.32   \\
              & MonoGS~\cite{monogs}        & 4.08$\pm$1.75 & 11.61$\pm$5.70    \\
Real-Virtual~ & AI-copilot~\cite{zhang2024ai}    & 6.39$\pm$2.07 & 14.03$\pm$6.12 \\
              & CycleGAN~\cite{cyclegan_}      & 6.05$\pm$3.35 & 13.29$\pm$6.54   \\
              & UNSB~\cite{unsb}          & 3.06$\pm$1.44 & 11.19$\pm$4.91   \\
              & EndoSERV(Ours)  & 1.90$\pm$0.51 & 6.22$\pm$2.83    
\end{tblr}
\end{table}

\begin{figure*}[!t]
\centering
\includegraphics[width=\linewidth]{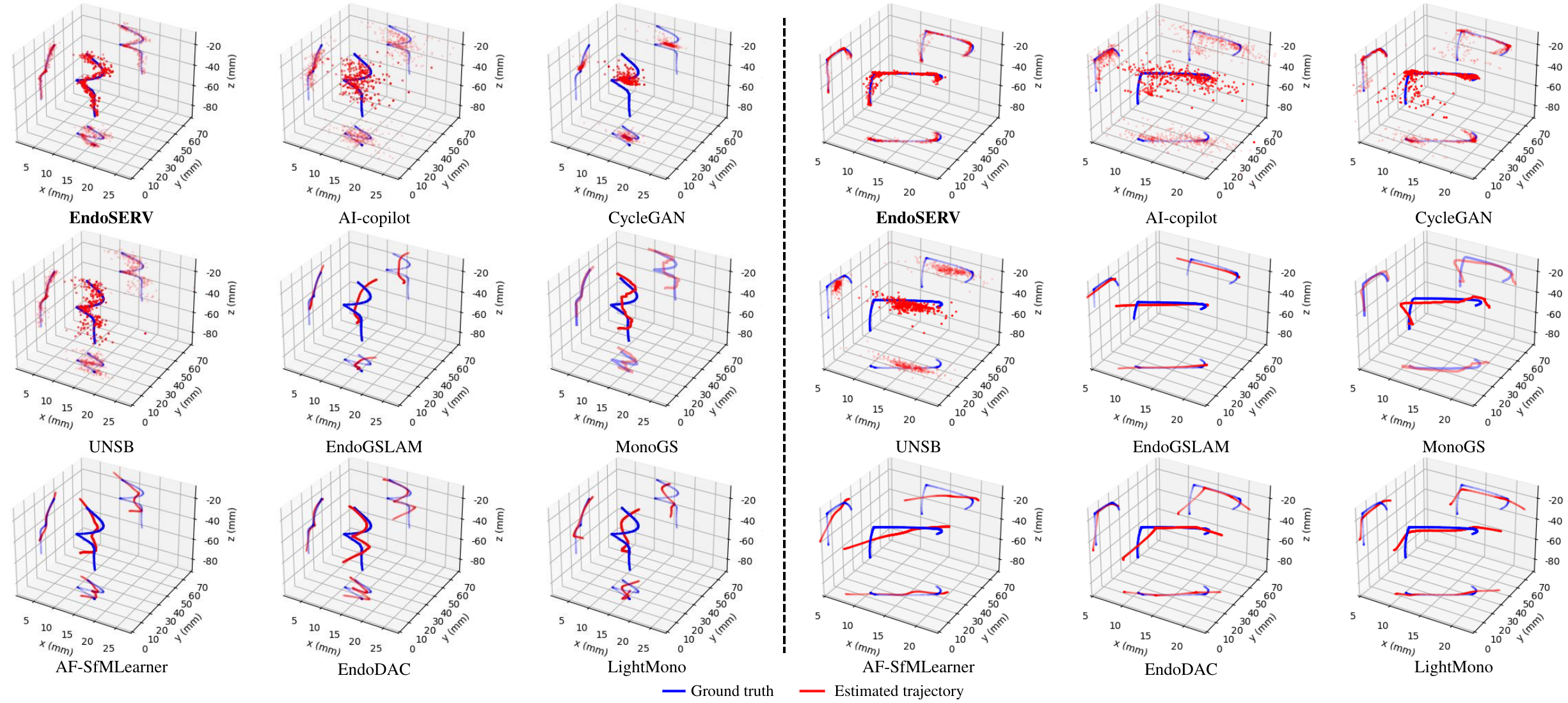}
\caption{ \textbf{Two examples of pose estimation from the C3VD dataset are provided. }  Visualizations for all methods are first centralized, followed by the presentation of 3D images along with their projections on the three coordinate axes.
}
\label{fig:c3vd_result} 
\end{figure*}

\begin{table}
\centering
\caption{$r_{RPE}$ results of Real-Virtual alignment methods}
\label{tab:r_rpe_results}
\begin{tblr}{
  cells = {c},
  cell{1}{1} = {r=2}{},    
  vline{2-3} = {1-6}{},  
  hline{1,3,7} = {-}{},  
  hline{2} = {2-3}{},    
  hline{6} = {1-3}{},    
}
Method                & C3VD dataset  & Clinical data     \\
                      & $r_{RPE} (deg)$     & $r_{RPE} (deg)$         \\
AI-copilot~\cite{zhang2024ai} & 2.72$\pm$0.56 & 4.16$\pm$1.93   \\
CycleGAN~\cite{cyclegan_}   & 1.87$\pm$0.83 & 3.14$\pm$1.72   \\
UNSB~\cite{unsb}            & 1.80$\pm$1.11 & 3.69$\pm$1.17   \\
EndoSERV(Ours)        & 0.50$\pm$0.26 & 1.05$\pm$0.44    
\end{tblr}
\end{table}

\section{Baseline Methods}

In our experiments, we compared eight baselines in three categories:
\subsection{Real-Virtual Alignment}
The real video frames are converted to the virtual domain for estimating the 6DoF pose of the endoscope. Multiple real-to-virtual transformation networks are adopted as the baselines.
\begin{itemize}
    \item CycleGAN~\cite{cyclegan_}: CycleGAN is an image-to-image translation method  in the absence of paired examples. It proposes an inverse mapping and a cycle consistency loss to achieve unpaired image translation.
    \item AI-copilot~\cite{zhang2024ai}: AI-copilot is a structure-preserving unpaired image translation method. It consists of a generator, a discriminator and a depth estimator, and leverages a depth constraint for structure consistency.
    \item UNSB~\cite{unsb}: UNSB expresses the Schrödinger Bridge problem as a sequence of adversarial learning problems and incorporates advanced discriminators and regularization to learn a Schrödinger Bridge between unpaired data.
\end{itemize}

\subsection{Neural Implicit-Based SLAM}
\begin{itemize}
    \item EndoGSLAM~\cite{endogslam}:  EndoGSLAM is an efficient SLAM approach for endoscopic surgeries, which integrates streamlined Gaussian representation and differentiable rasterization to facilitate online camera tracking and tissue reconstructing.
    \item MonoGS~\cite{monogs}: MonoGS is the first application of 3D Gaussian Splatting (3DGS) in monocular SLAM. It formulates camera tracking for 3DGS using direct optimisation against the 3D Gaussians, and introduce geometric verification and regularization to handle the ambiguities occurring in incremental 3D dense reconstruction.
\end{itemize}

\subsection{Structure from Motion}
\begin{itemize}
    \item AF-SfMLearner~\cite{AF}: AF-SfMLearner is a self-supervised framework to estimate monocular depth and ego motion simultaneously in endoscopic scenes. It introduces a novel concept referred to as appearance flow to address the brightness inconsistency problem.
    \item EndoDAC~\cite{endodac}: EndoDAC is an efficient self-supervised depth estimation framework that adapts foundation models to endoscopic scenes. It develops the Dynamic Vector-Based Low-Rank Adaptation (DV-LoRA) and employ Convolutional Neck blocks to tailor the foundational model to the surgical domain, utilizing remarkably few trainable parameters.
    \item LightMono~\cite{zhang2023lite}: LightMono is a hybrid architecture combined with CNNs and Transformers. It  extracts rich multi-scale local features, and  takes advantage of the self-attention mechanism to encode long range global information into the features.

\end{itemize}

\section{Experiments with Simple Dataset}
\subsection{Experiment Settings}
We evaluated the proposed method using the C3VD dataset \cite{c3vd}, which consists of 22 colonoscopic video sequences paired with corresponding virtual models. Due to the small size and relatively low complexity of the C3VD dataset, we trained the model using only 50 real images per case to assess its generalization ability.

\subsection{Training Implementations}

During the offline training phase, the prompts we used are:``Bleed, inflamed red, slightly reddish, pale yellow and orange areas, pinkish hue, ulcerated regions, a sketch with crayon, mosaic.''  The pretrained diffusion model we used is InstructPix2Pix \cite{instructpix2pix}. For the feature encoder and the scene coordinate head, we follow previous work DSAC* \cite{dsac} and applied the pretrained feature encoder in ACE \cite{ace} as the initialization. We used Adam as the optimizer, with the learning rate 1e-4 and the training iteration 1e+6. We used CycleGAN as the foundation style transfer model, which trains for 200 epochs using 50 real images and all virtual images per case as the training data. The testing images include all real images. The learning rate is 2e-4 and the optimizer is Adam. During the online phase, since the simple dataset can be considered as a single sub-segment, the retrieval phase is not needed. In addition, a single transfer is enough to cover the whole scene; therefore, the number of the sliding window is one. The experiments on the simple dataset focus on testing the efficiency of offline training.

\subsection{Results}

To assess the performance of our algorithm, we conducted experiments on the C3VD dataset, treating each video segment as a separate sub-segment due to the limited number of frames in each video. As shown in Table \ref{tab:synthetic}, EndoSERV consistently outperforms all baselines in terms of Absolute Trajectory Error (ATE), achieving the lowest ATE of 1.90 $\pm$ 0.51 $mm$.

SfM-based methods, including AF-SfMLearner, EndoDAC, and LightMono, demonstrate ATE values ranging from 3.89 $mm$ to 4.81 $mm$, with EndoDAC achieving the best performance, 3.89 $\pm$ 2.50 $mm$. 
MonoGS achieves the lowest ATE (4.08 $\pm$ 1.75 $mm$) among NeRF-SLAM methods, demonstrating its effectiveness in leveraging neural radiance fields for pose estimation. 
However, EndoSERV outperforms all SfM and NeRF-SLAM methods, even without 7Dof-alignment. We conduct the t-test between EndoSERV and EndoDAC; the p-value is 3.512e-5, demonstrating the improvement is statistically significant.

Real-virtual alignment methods, which overcome the need of 7Dof-alignment with pose ground truth, exhibit competing performance on ATE evaluation.
AI-copilot and CycleGAN report the highest ATE values (6.39 $\pm$ 2.07 $mm$ and 6.05 $\pm$ 3.35 $mm$, respectively), while UNSB perform better with an ATE of 3.06 $\pm$ 1.44 $mm$. Despite these results, EndoSERV outperforms all Real-Virtual Alignment methods, achieving significant improvement in ATE over the best-performing real-virtual alignment method, UNSB. Additionally, EndoSERV achieved the best $r_{RPE}$ of 0.50$\pm$0.26 $deg$, outperforming all competing real-virtual alignment methods.

Addition, the trajectory visualization is shown in Fig. \ref{fig:c3vd_result}, providing qualitative insights into the performance of different approaches. Notably, our proposed method, EndoSERV, demonstrates significant advantages compared to existing methods.

\section{Experiments with Robotic System}
\begin{figure*}[!t]
\centering
\includegraphics[width=\linewidth]{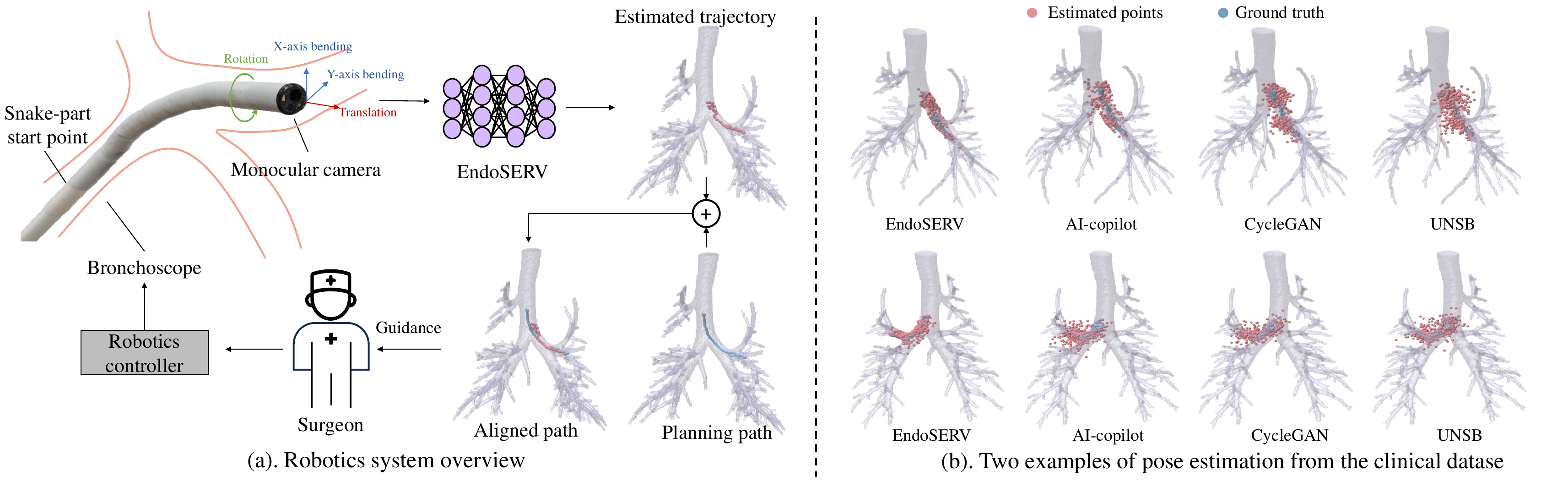}
\caption{ \textbf{(a). Robotics system overview.} The robotic system is equipped with the continuum joints allowing 4DoF operations including translation, rotation and X/Y bending. During surgery, the surgeon controls the robot system, capturing intra-operative images from the monocular camera. EndoSERV module estimates the absolute pose and generates the trajectory for intra-operative guidance. \textbf{(b). Two examples of pose estimation from the clinical dataset are provided.} All Real-Virtual Alignment Methods overcome the need of 7-DoF alignment with pose ground truth. EndoSERV achieves the best performance, whose trajectory crosses through the airway less frequently.
}

\label{fig:robotics} 
\end{figure*}

\subsection{Experiment Settings}

We implemented the localization system on an endobronchial surgical robot and conducted the in-vivo animal trial. The study was conducted at our collaboration hospital with ethical approval by the IRB\footnote{The names of company, hospital and the ethical approval number are anonymized due to the double-blinded policy by IEEE T-RO. We will publish these details if the manuscript is accepted.}. All procedures were performed in accordance with ethical standards. 

As shown in Fig.\ref{fig:robotics}, the robotic system is equipped with the continuum joints allowing 4DoF operations including translation, rotation, and X/Y bending. The outer diameter of the continuum joints is 4.2$mm$. Both \textit{in-vivo} images and their corresponding virtual images were acquired using a robotics platform. For the virtual images, we imported segmented bronchial trees from pre-operative CT scans into Unity and rendered the virtual images. For the \textit{in-vivo} images, we acquired 52,340 endoscopy frames from 6 pigs, including 33 video sequences. In-vivo images were captured during the animal trial with a bronchoscope operating at 25 frames per second (fps).
For the pose ground truth in \textit{in-vivo} datasets, we first applied a coarse relative pose estimation algorithm to obtain the coarse pose, and refined it by  manually aligning the structure between real images and virtual renderings.
Specifically, we imported the initial estimated poses, along with the segmented bronchial tree, into a virtual engine for rendering. If the pose was accurate, the rendered virtual image should align with the real image in terms of structure. If misalignment was detected, we manually adjusted the pose within the virtual engine, iterating until the virtual image and the real image structures nearly overlapped.

To ensure a fair evaluation across different pose estimation methods, we adopt different initialization and alignment strategies tailored to the nature of each method. For SfM-based and neural implicit-based SLAM approaches, which inherently produce relative pose estimates between consecutive frames, we initialize the first frame’s pose at the origin of the coordinate system. Subsequent poses are then accumulated using a SLAM-based pipeline to form a complete trajectory. During evaluation, we apply a 7-DoF transformation to align the estimated trajectory with the ground truth. In contrast, for real-virtual alignment methods, the availability of virtual renderings with known absolute poses enables direct prediction of the endoscope’s global pose. This obviates the need for post-hoc alignment, as the pose predictions are already in the same coordinate space as the ground truth.

During the operation, as the surgeon manipulates the robot, the bronchoscope's monocular camera captures intra-operative images, which are processed by the EndoSERV module to estimate the bronchoscope’s absolute pose and generate a trajectory. This trajectory is compared with the pre-operative planned path, providing real-time guidance to the surgeon and helping identify deviations for correction.

\subsection{Training Implementations}
The offline training setting is the same as the simple dataset. We used CycleGAN as the foundation style transfer model, which trains for 200 epochs using 20 sequences as the training data. The learning rate is 2e-4 and the optimizer is Adam. 
During the online training phase, we first used R2former \cite{zhu2023r2former} as the feature extractor for the retrieval step. The size of real buffer and virtual buffer is 100. For the transfer fine-tuning and deformation refinement, we can use two GPUs to train them in parallel. Specifically, one GPU fine-tunes the CycleGAN model for only 8 epochs. Another GPU runs inference on the CycleGAN model and feeds generated images to the `Transfer refinement' module and `Odometry training' module.
The refinement decoder trains for 20 epochs with a learning rate of 2e-5, with a batch size of 12. All experiments are conducted on GPUs with NVIDIA GeForce RTX 3090.

\subsection{Results}

In this section, we conducted experiments on the clinical data, which are more challenging due to the complex structure and unpredictable artifacts.

The evaluation of camera absolute pose estimation methods on the clinical dataset was conducted using the Absolute Trajectory Error (ATE). The results, summarized in Table \ref{tab:synthetic}, demonstrate that our proposed method, EndoSERV, outperforms all other approaches, achieving the lowest ATE of 6.22 $\pm$ 2.83 $mm$.

Among the SfM methods, AF-SfMLearner and LightMono demonstrate similar performance with ATE values of 12.49 $\pm$ 7.56 mm and 12.26 $\pm$ 5.97 $mm$, respectively, while EndoDAC achieves a slightly lower error of 11.10 $\pm$ 5.63 $mm$. NeRF-SLAM-based methods such as EndoGSLAM and MonoGS report ATE of 12.04 $\pm$ 5.32 $mm$ and 11.61 $\pm$ 5.70 $mm$, respectively, showing comparable performance to the SfM approaches but trailing behind EndoSERV.

Real-Virtual Alignment methods exhibit variable performance on the clinical dataset. AI-copilot and CycleGAN report ATE values of 14.03 $\pm$ 6.12 $mm$ and 13.29 $\pm$ 6.54 $mm$, respectively, indicating higher errors compared to SfM and NeRF-SLAM methods. UNSB demonstrates a more competitive performance with an ATE of 11.19 $\pm$ 4.91 $mm$, demonstrating the strong ability of the diffusion-based model. For the rotation metric, $r_{RPE}$, EndoSERV achieves an error of 1.05$\pm$0.44 $deg$. This result substantially outperforms all baselines: AI-copilot (4.16$\pm$1.93), CycleGAN (3.14$\pm$1.72), and UNSB (3.69$\pm$1.17).

In addition, the trajectory visualization is shown in Fig.\ref{fig:robotics}, providing qualitative insights into the performance of different approaches. Notably, our proposed method, EndoSERV, demonstrates significant advantages compared to existing methods.
Real-Virtual Alignment baselines, including AI-copilot, CycleGAN, and UNSB, estimate absolute camera poses directly, eliminating the need for 7-DoF alignment during testing.
However, they tend to exhibit more dispersed trajectories due to the frame-wise pose estimation approach, leading to discontinuities in the reconstructed path.

Our proposed method, EndoSERV, overcomes these limitations by proposing the texture-agnostic offline pretraining and refining the distortion and deformation. 
As a real-virtual alignment method, EndoSERV obviates the need for 7-DOF alignment and real-world labels, thereby ensuring its applicability in realistic surgical scenarios. At the same time, EndoSERV achieves a trajectory that is more continuous than other real-virtual alignment baselines, with less frequency crossing through the airway model, highlighting its robustness and accuracy.

\section{Discussion}

\begin{figure*}[!t]
\centering
\includegraphics[width=\linewidth]{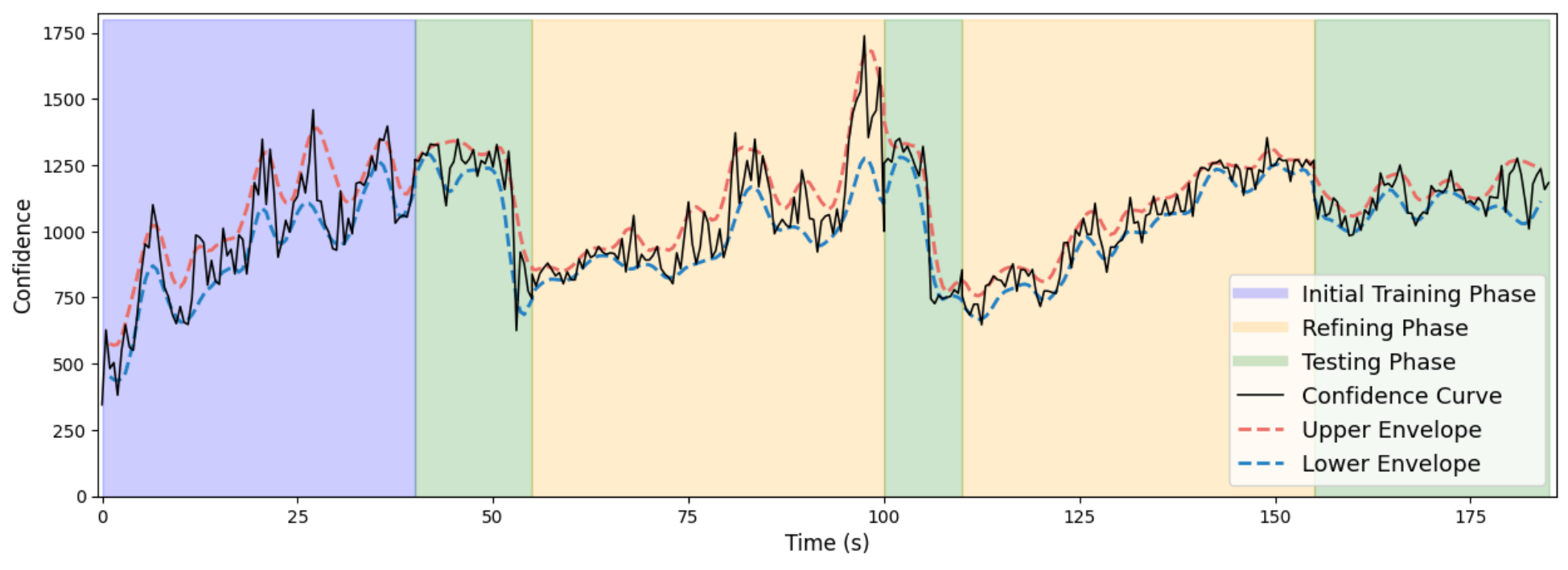}
\caption{ \textbf{Surgical Guidance with Confidence-Aware Camera Localization.} The process is composed of two refining phases and three testing phases. During the refining phase, the confidence progressively increases, while in the testing phase, the confidence experiences a decline at a specific time point.
}
\label{fig:sequence_time} 
\end{figure*}
\subsection{Confidence-Aware Surgical Guidance Analysis}

In this section, we present a specific case from the clinical data to demonstrate how EndoSERV estimates camera pose accurately and efficiently.
As shown in Fig. \ref{fig:sequence_time}, the process begins with a surgeon performing a multi-angle scan at the entry point to train an initial model for localization. After the initial training, the system transitions to testing, achieving real-time performance (27 fps). Testing proceeds until a significant drop in the confidence signal is detected, signaling the need for further refining.
The refining phase, lasting approximately 45 seconds, can run parallel with the testing phase. It fine-tunes the transfer model, refinement decoder, and the scene coordinate head, improving the estimating confidence. Upon completion of this refinement, the system reverts to the testing phase, and this cycle repeats.

In the example shown in Fig. \ref{fig:sequence_time}, a dataset of 3,000 images required two instances of online refinement. Except for the initial training time, this system only required 143 seconds to localize 3,000 images, achieving efficient surgical guidance for surgeons. The inference time of each module is shown in Table \ref{tab:time}, demonstrating the efficiency.

\begin{table}[]
\centering
\caption{Computational Time Evaluation}
\label{tab:time}
\begin{tblr}{
  cells = {c},
  cell{1}{1} = {c=2,r=1}{},
  vline{2} = {1-6}{},
  colspec = {c X[1,c] c}, 
  hline{1,2,6,7} = {-}{},
}
Different modules in EndoSERV        &  Result  \\
Style Transfer module  &  6.70ms \\
Refinement Module  &  15.60ms \\
Scene Coordinate Regression  &  2.94ms \\
PnP \& RANSAC  &  12.48ms \\
Total time & 37.73ms \\
\end{tblr}
\end{table}

\subsection{Impact of Each Module: Feature Similarity Analysis}

In this section, we evaluate the contributions of individual components by examining their ability to enhance the similarity between virtual and real features. This capability is crucial, as it enables the pose estimation network to be trained in the virtual domain while generalizing effectively to real-world scenarios.
We hypothesize that two primary factors influence this feature similarity: (1) the effectiveness of the Real-Virtual Alignment strategy used to bridge the domain gap, and (2) at the feature level, the feature encoder's ability to extract texture-agnostic features.

To investigate these factors, our experiments use real images and their corresponding virtual counterparts from the clinical dataset. We assess two distinct encoder configurations to evaluate their feature extraction capabilities: (1). ACE-encoder: A feature encoder pretrained on large datasets \cite{ace}. (2). Offline-encoder: The encoder developed and pretrained during the offline pretraining phase of this work.

Furthermore, we compare four different real-virtual alignment strategies to measure their alignment effectiveness: (1) None (original real image), (2) EndoSERV, (3) AI-copilot, and (4) UNSB. Specifically, the `EndoSERV' strategy listed here refers to a sub-component of our proposed method, which consists of a feature encoder and a refinement decoder to generate virtual images.

For each real-virtual image pair, the real image is processed by one of the four real-virtual alignment strategies (or `None'). Then, both the processed real image and the original virtual image are passed through one of the two feature encoders (ACE-encoder or Offline-encoder) to extract their respective feature vectors. We then compute the Cosine Similarity between these two vectors. A higher cosine similarity score indicates a smaller domain gap and better alignment at the feature level.

Fig. \ref{fig:sim} summarizes the similarity scores: 

\begin{figure}[]
\centering
\includegraphics[width=\linewidth]{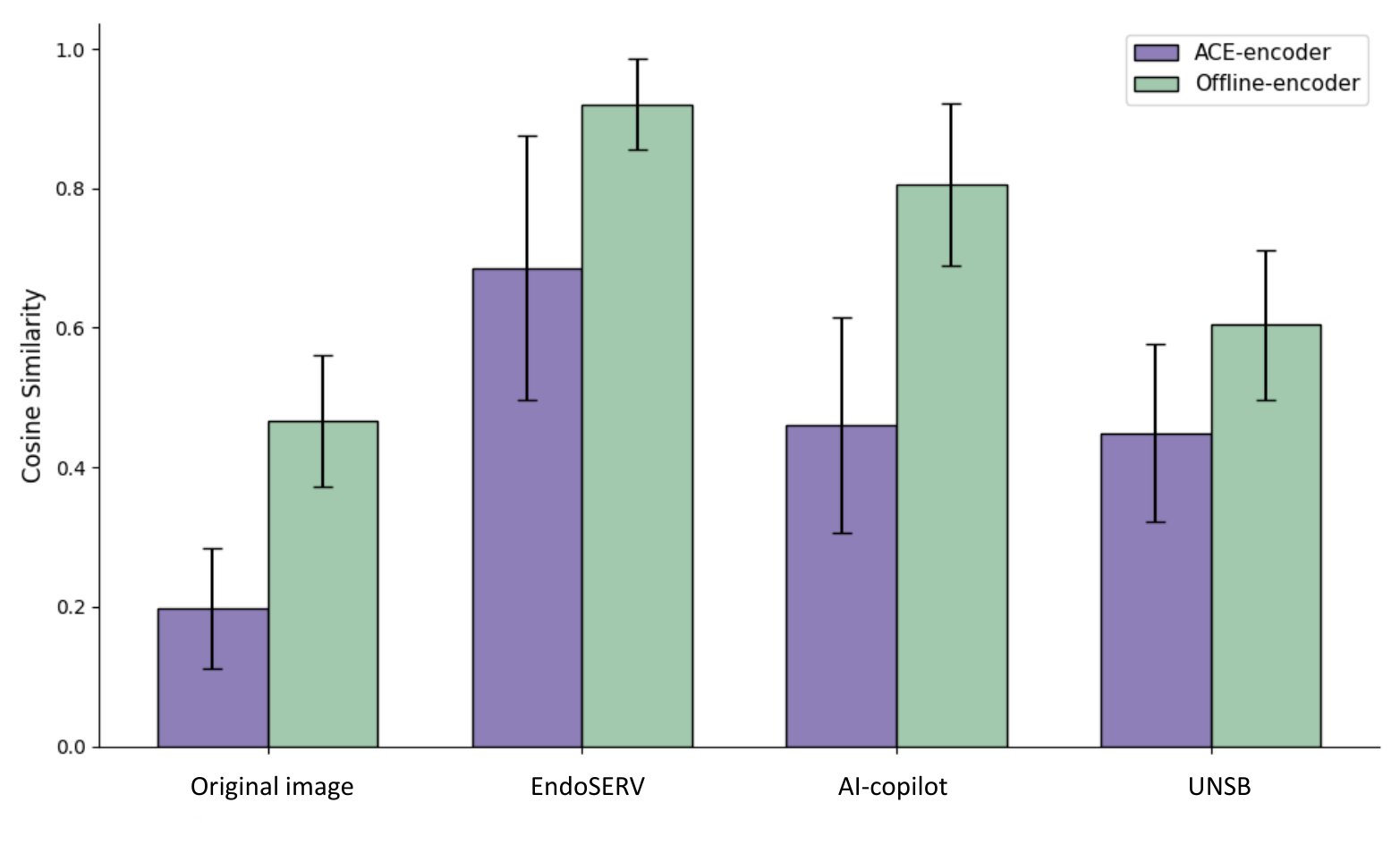}
\caption{ \textbf{Real-to-Virtual feature similarity for different configurations.} It measures the feature similarity using different feature encoders and Real-Virtual Alignment methods. The similarity is measured using cosine similarity. 
}
\label{fig:sim} 
\end{figure}

\textbf{Comparison Across Real-Virtual Alignment Strategies:}
The EndoSERV strategy consistently achieved the highest similarity scores across encoder configurations, demonstrating its effectiveness in bridging the domain gap. For the ACE-encoder, the EndoSERV strategy improved similarity to 0.6856 $\pm$ 0.1897, significantly outperforming the original image (0.1973 $\pm$ 0.0858), AI-copilot (0.4569 $\pm$ 0.1678), and UNSB (0.4605 $\pm$ 0.1537). A similar trend was observed with the Offline-encoder, where EndoSERV yielded a similarity score of 0.9208 $\pm$ 0.0660, surpassing AI-copilot (0.6113 $\pm$ 0.1413) and UNSB (0.8059 $\pm$ 0.1169). These findings confirm that EndoSERV is a more effective adaptation method than other strategies, as it employs a paired generation strategy to generate virtual images that are more closely aligned with the virtual domain.

\begin{figure*}[]
\centering
\includegraphics[width=\linewidth]{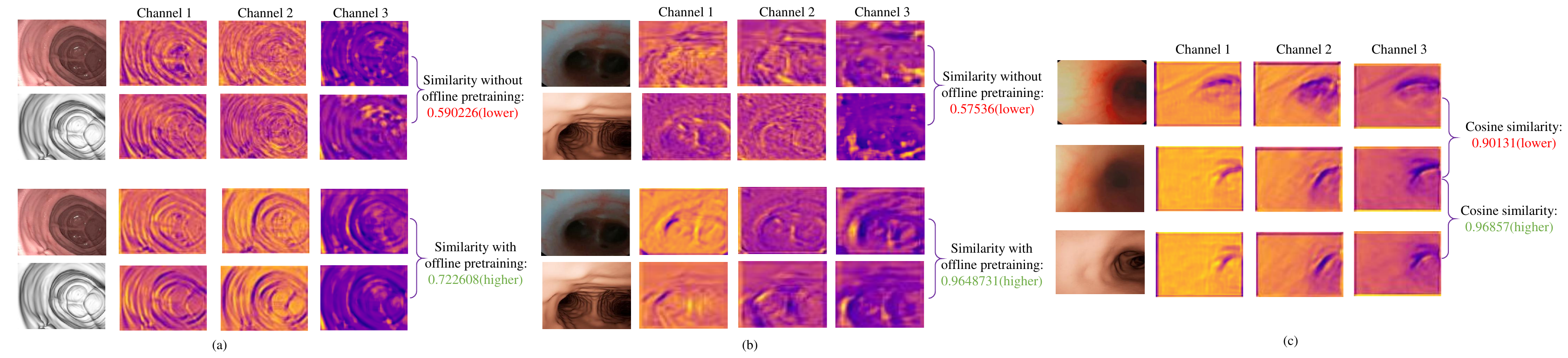}
\caption{\textbf{Texture-agnostic features for real-virtual domain gap and discriminative features for appearance-similar endoscopic scenarios}. (a) and (b) show the feature maps extracted from virtual and real images on the C3VD dataset and the clinical dataset, respectively, with and without offline pretraining. It can be observed that offline pretraining significantly increases the similarity between the extracted virtual and real features. (c) illustrates that although images from different bronchial branches may exhibit anatomical similarity, EndoSERV is still capable of extracting distinctive features. Specifically, for images from the same location but different domains, EndoSERV extracts highly similar features. In contrast, for anatomically similar images from different locations, it is able to capture distinct features.}
\label{fig:all_feature_map} 
\end{figure*}

\textbf{Impact Across Encoder Configurations:} For different image encoders, 
the Offline-encoder consistently outperformed the ACE-encoder across all adaptation strategies, demonstrating the advantages of offline training. For instance, under the EndoSERV strategy, the Offline-encoder achieved a similarity score of 0.9208 $\pm$ 0.0660, compared to 0.6856 $\pm$ 0.1897 for the ACE-encoder. Similarly, for UNSB, the Offline-encoder achieved 0.8059 $\pm$ 0.1169, significantly higher than 0.4605 $\pm$ 0.1537 for the ACE-encoder. 
These results demonstrate the Offline-encoder's superior ability to reduce the domain gap, especially when combined with advanced adaptation techniques.

\subsection{Ablation study of augmentation strategies}

In EndoSERV, two augmentation strategies are essential: the texture-diverse augmentation in the offline training phase, and the distortion and deformation augmentation during the online training phase. 
In this section, we investigate the impact of each augmentation strategy.

\begin{table}[]
    \centering
    \caption{Ablation Study on each augmentation strategy in EndoSERV}
    \resizebox{0.6\linewidth}{!}{
    \begin{tabular}{c|c|c}
        \hline
        \multicolumn{2}{c|}{Augmentations} & \multirow{2}{*}{ATE} \\
        \cline{1-2}
        Texture diverse & DDAug & \\
        \hline
        \xmark & \xmark & 9.03$\pm$4.68\\
        \cmark & \xmark & 8.16$\pm$4.37\\
        \xmark & \cmark & 8.80$\pm$4.69\\
        \cmark & \cmark & \textbf{6.22$\pm$2.83}\\
        \hline
        \multicolumn{2}{c|}{Rand-Aug} & 7.59$\pm$4.31\\
        \multicolumn{2}{c|}{Auto-Aug} & 7.18$\pm$3.23\\
        \hline
    \end{tabular}
    }
    \label{tab:ablation}
\end{table}

To evaluate the effectiveness of different augmentation strategies in enhancing the performance of EndoSERV, we conducted an ablation study focusing on two specific augmentation techniques: Texture Diverse and DDAug. In the absence of Texture Diverse augmentation, the feature encoder was trained exclusively on virtual images. Similarly, without DDAug, the refine decoder reconstructed virtual images using only the transferred real images.

Table \ref{tab:ablation} presents the results of this study. The baseline model, without any augmentation applied, achieved an ATE of 9.03$\pm$4.68 $mm$. Introducing the Texture Diverse augmentation alone resulted in a significant reduction of the ATE to 8.16$\pm$4.37 $mm$, demonstrating its substantial contribution to improving localization accuracy. Applying DDAug in isolation yielded a decrease in ATE to 8.80$\pm$4.69 $mm$.
Notably, the combination of both Texture Diverse and DDAug augmentations led to the most pronounced improvement, reducing the ATE to 6.22$\pm$2.83 $mm$. This synergistic effect underscores the complementary nature of the two augmentation strategies, where Texture Diverse primarily enhances the model's ability to generalize across varied textures, and DDAug simulates the distortion and deformation scenarios, fostering a more robust localization capability.

Additionally, for simulating the real endoscopic scenarios, we carefully design the augmentation strategy. To validate it, we compare some augmentation methods usually used in deep learning, like Rand-Aug and Auto-Aug. The results are shown in Table \ref{tab:ablation}. Using data augmentation techniques such as Rand-Aug or Auto-Aug results in a lower ATE compared to training without any augmentation. However, their performance is still inferior to that achieved with our proposed DDAug. This improvement can be attributed to DDAug’s ability to more realistically simulate the challenges encountered in endoscopic scenarios, thereby enhancing the robustness of the network.

\subsection{Texture-agnostic feature and similar anatomical feature}
In this section, we investigate two common challenges in endoscopic image processing from the perspective of feature maps. First, we examine whether the feature extractor obtained through offline training can learn texture-agnostic representations. Second, we assess whether the extractor can generate distinguishable features for images captured from different branches that exhibit nearly identical geometric and topological properties, thereby enabling accurate localization.\\
The results are shown in Fig. \ref{fig:all_feature_map}. For the feature extractor without pretraining, there is a substantial discrepancy between the feature maps of virtual and real images. However, after offline pretraining, this gap is significantly reduced. In Fig. \ref{fig:all_feature_map} (c), we select two images from different bronchial branches that share a highly similar appearance. It can be observed that our feature extractor is still able to produce distinct features, demonstrating its ability to differentiate between visually similar yet anatomically distinct regions.

\subsection{Effectiveness of estimated confidence-based filtering}
To further evaluate the effectiveness of our confidence estimation, we incorporate an effective post-processing step that leverages the confidence estimated from scene coordinate predictions. This confidence-aware refinement filters out predictions with low confidence, which are often associated with outliers and temporal inconsistencies. As shown in Table~\ref{tab:uncertainty_filtering_}, the proposed confidence-based filtering significantly improves the accuracy and stability of pose estimation. Starting from a baseline ATE of $6.22 \pm 2.83$ without any filtering, the ATE consistently decreases to $5.81 \pm 2.50$, $5.58 \pm 2.41$, $5.44 \pm 2.26$, and $5.37 \pm 2.19$, when filtering out the lowest 10\%, 20\%, 30\% and 40\% of confident predictions, respectively. This reduction in ATE demonstrates that the estimated confidence effectively captures prediction reliability and can be used to suppress noise and outliers.

\begin{table}[ht]
\centering
\caption{Effect of Uncertainty-Based Filtering on Pose Estimation Accuracy (ATE).}
\resizebox{0.8\linewidth}{!}{
\begin{tabular}{cc}
\hline
\textbf{Uncertainty Filter Threshold} & \textbf{ATE (↓)}\\
\hline
No Filter & $6.22 \pm 2.83$ \\
Top 90\% confidence (10\% filtered) & $5.81 \pm 2.50$  \\
Top 80\% confidence (20\% filtered) & $5.58 \pm 2.41$  \\
Top 70\% confidence (30\% filtered) & $5.44 \pm 2.26$  \\
Top 60\% confidence (40\% filtered) & $5.37 \pm 2.19$  \\
\hline
\end{tabular}
}
\label{tab:uncertainty_filtering_}
\end{table}

\section{Conclusion}

In this paper, we propose EndoSERV, a novel approach for efficient endoluminal robot localization. 
We propose an adaptive segment-to-structure strategy that partitions extended luminal paths into manageable sub-segments. For each sub-segment, we map everything in the real domain to the pre-operative virtual domain, taking advantage of virtual ground truth for odometry training. 
Specifically, we design a robust offline pretraining to extract texture-agnostic features, and propose a fast online training for simultaneous domain adaptation and odometry training.
Additionally, we propose a novel augmentation-then-recovery strategy, simulating potential distortion and deformation in real scenarios and recovering them via a paired training pipeline. 
Experimental results on both public and clinical datasets demonstrate significantly improved performance over current state-of-the-art methods, even in the absence of real-world pose labels.
\bibliographystyle{IEEEtran}
\bibliography{refs}

@article{sfm,
  title={Evaluation and stability analysis of video-based navigation system for functional endoscopic sinus surgery on in vivo clinical data},
  author={Leonard, Simon and Sinha, Ayushi and Reiter, Austin and Ishii, Masaru and Gallia, Gary L and Taylor, Russell H and Hager, Gregory D},
  journal={IEEE transactions on medical imaging},
  volume={37},
  number={10},
  pages={2185--2195},
  year={2018},
  publisher={IEEE}
}

@article{AF,
  title={Self-supervised monocular depth and ego-motion estimation in endoscopy: Appearance flow to the rescue},
  author={Shao, Shuwei and Pei, Zhongcai and Chen, Weihai and Zhu, Wentao and Wu, Xingming and Sun, Dianmin and Zhang, Baochang},
  journal={Medical image analysis},
  volume={77},
  pages={102338},
  year={2022},
  publisher={Elsevier}
}

@article{endoslam,
  title={EndoSLAM dataset and an unsupervised monocular visual odometry and depth estimation approach for endoscopic videos},
  author={Ozyoruk, Kutsev Bengisu and Gokceler, Guliz Irem and Bobrow, Taylor L and Coskun, Gulfize and Incetan, Kagan and Almalioglu, Yasin and Mahmood, Faisal and Curto, Eva and Perdigoto, Luis and Oliveira, Marina and others},
  journal={Medical image analysis},
  volume={71},
  pages={102058},
  year={2021},
  publisher={Elsevier}
}

@article{surgicaldino,
  title={Surgical-DINO: adapter learning of foundation models for depth estimation in endoscopic surgery},
  author={Cui, Beilei and Islam, Mobarakol and Bai, Long and Ren, Hongliang},
  journal={International Journal of Computer Assisted Radiology and Surgery},
  pages={1--8},
  year={2024},
  publisher={Springer}
}

@article{luo2014discriminative,
  title={A discriminative structural similarity measure and its application to video-volume registration for endoscope three-dimensional motion tracking},
  author={Luo, Xiongbiao and Mori, Kensaku},
  journal={IEEE transactions on medical imaging},
  volume={33},
  number={6},
  pages={1248--1261},
  year={2014},
  publisher={IEEE}
}

@article{luo2012development,
  title={Development and comparison of new hybrid motion tracking for bronchoscopic navigation},
  author={Lu{\'o}, Xi{\'o}ngbi{\=a}o and Feuerstein, Marco and Deguchi, Daisuke and Kitasaka, Takayuki and Takabatake, Hirotsugu and Mori, Kensaku},
  journal={Medical image analysis},
  volume={16},
  number={3},
  pages={577--596},
  year={2012},
  publisher={Elsevier}
}

@article{zhang2024ai,
  title={AI co-pilot bronchoscope robot},
  author={Zhang, Jingyu and Liu, Lilu and Xiang, Pingyu and Fang, Qin and Nie, Xiuping and Ma, Honghai and Hu, Jian and Xiong, Rong and Wang, Yue and Lu, Haojian},
  journal={Nature communications},
  volume={15},
  number={1},
  pages={241},
  year={2024},
  publisher={Nature Publishing Group UK London}
}

@inproceedings{endodac,
  title={Endodac: Efficient adapting foundation model for self-supervised depth estimation from any endoscopic camera},
  author={Cui, Beilei and Islam, Mobarakol and Bai, Long and Wang, An and Ren, Hongliang},
  booktitle={International Conference on Medical Image Computing and Computer-Assisted Intervention},
  pages={208--218},
  year={2024},
  organization={Springer}
}

@inproceedings{tcl,
  title={TCL: Triplet Consistent Learning for Odometry Estimation of Monocular Endoscope},
  author={Yue, Hao and Gu, Yun},
  booktitle={International Conference on Medical Image Computing and Computer-Assisted Intervention},
  pages={144--153},
  year={2023},
  organization={Springer}
}

@article{liu2019dense,
  title={Dense depth estimation in monocular endoscopy with self-supervised learning methods},
  author={Liu, Xingtong and Sinha, Ayushi and Ishii, Masaru and Hager, Gregory D and Reiter, Austin and Taylor, Russell H and Unberath, Mathias},
  journal={IEEE transactions on medical imaging},
  volume={39},
  number={5},
  pages={1438--1447},
  year={2019},
  publisher={IEEE}
}

@inproceedings{imap,
  title={imap: Implicit mapping and positioning in real-time},
  author={Sucar, Edgar and Liu, Shikun and Ortiz, Joseph and Davison, Andrew J},
  booktitle={Proceedings of the IEEE/CVF international conference on computer vision},
  pages={6229--6238},
  year={2021}
}

@inproceedings{inerf,
  title={inerf: Inverting neural radiance fields for pose estimation},
  author={Yen-Chen, Lin and Florence, Pete and Barron, Jonathan T and Rodriguez, Alberto and Isola, Phillip and Lin, Tsung-Yi},
  booktitle={2021 IEEE/RSJ International Conference on Intelligent Robots and Systems (IROS)},
  pages={1323--1330},
  year={2021},
  organization={IEEE}
}

@article{enerf,
  title={ENeRF-SLAM: A Dense Endoscopic SLAM With Neural Implicit Representation},
  author={Shan, Jiwei and Li, Yirui and Xie, Ting and Wang, Hesheng},
  journal={IEEE Transactions on Medical Robotics and Bionics},
  year={2024},
  publisher={IEEE}
}

@inproceedings{endogslam,
  title={Endogslam: Real-time dense reconstruction and tracking in endoscopic surgeries using gaussian splatting},
  author={Wang, Kailing and Yang, Chen and Wang, Yuehao and Li, Sikuang and Wang, Yan and Dou, Qi and Yang, Xiaokang and Shen, Wei},
  booktitle={International Conference on Medical Image Computing and Computer-Assisted Intervention},
  pages={219--229},
  year={2024},
  organization={Springer}
}

@inproceedings{eslam,
  title={Eslam: Efficient dense slam system based on hybrid representation of signed distance fields},
  author={Johari, Mohammad Mahdi and Carta, Camilla and Fleuret, Fran{\c{c}}ois},
  booktitle={Proceedings of the IEEE/CVF Conference on Computer Vision and Pattern Recognition},
  pages={17408--17419},
  year={2023}
}

@inproceedings{coslam,
  title={Co-slam: Joint coordinate and sparse parametric encodings for neural real-time slam},
  author={Wang, Hengyi and Wang, Jingwen and Agapito, Lourdes},
  booktitle={Proceedings of the IEEE/CVF Conference on Computer Vision and Pattern Recognition},
  pages={13293--13302},
  year={2023}
}

@inproceedings{nicerslam,
  title={Nicer-slam: Neural implicit scene encoding for rgb slam},
  author={Zhu, Zihan and Peng, Songyou and Larsson, Viktor and Cui, Zhaopeng and Oswald, Martin R and Geiger, Andreas and Pollefeys, Marc},
  booktitle={2024 International Conference on 3D Vision (3DV)},
  pages={42--52},
  year={2024},
  organization={IEEE}
}

@article{densenerfslam,
  title={Dense rgb slam with neural implicit maps},
  author={Li, Heng and Gu, Xiaodong and Yuan, Weihao and Yang, Luwei and Dong, Zilong and Tan, Ping},
  journal={arXiv preprint arXiv:2301.08930},
  year={2023}
}

@inproceedings{cyclegan_,
  title={Unpaired image-to-image translation using cycle-consistent adversarial networks},
  author={Zhu, Jun-Yan and Park, Taesung and Isola, Phillip and Efros, Alexei A},
  booktitle={Proceedings of the IEEE international conference on computer vision},
  pages={2223--2232},
  year={2017}
}

@inproceedings{cut,
  title={Contrastive learning for unpaired image-to-image translation},
  author={Park, Taesung and Efros, Alexei A and Zhang, Richard and Zhu, Jun-Yan},
  booktitle={Computer Vision--ECCV 2020: 16th European Conference, Glasgow, UK, August 23--28, 2020, Proceedings, Part IX 16},
  pages={319--345},
  year={2020},
  organization={Springer}
}

@article{diffusion,
  title={Denoising diffusion probabilistic models},
  author={Ho, Jonathan and Jain, Ajay and Abbeel, Pieter},
  journal={Advances in neural information processing systems},
  volume={33},
  pages={6840--6851},
  year={2020}
}

@article{unit-ddpm,
  title={Unit-ddpm: Unpaired image translation with denoising diffusion probabilistic models},
  author={Sasaki, Hiroshi and Willcocks, Chris G and Breckon, Toby P},
  journal={arXiv preprint arXiv:2104.05358},
  year={2021}
}

@article{EGSDE,
  title={Egsde: Unpaired image-to-image translation via energy-guided stochastic differential equations},
  author={Zhao, Min and Bao, Fan and Li, Chongxuan and Zhu, Jun},
  journal={Advances in Neural Information Processing Systems},
  volume={35},
  pages={3609--3623},
  year={2022}
}

@article{unsb,
  title={Unpaired Image-to-Image Translation via Neural Schr$\backslash$" odinger Bridge},
  author={Kim, Beomsu and Kwon, Gihyun and Kim, Kwanyoung and Ye, Jong Chul},
  journal={arXiv preprint arXiv:2305.15086},
  year={2023}
}

@inproceedings{augment,
  title={Augmenting colonoscopy using extended and directional cyclegan for lossy image translation},
  author={Mathew, Shawn and Nadeem, Saad and Kumari, Sruti and Kaufman, Arie},
  booktitle={Proceedings of the IEEE/CVF Conference on Computer Vision and Pattern Recognition},
  pages={4696--4705},
  year={2020}
}

@inproceedings{pix2pix,
  title={Image-to-image translation with conditional adversarial networks},
  author={Isola, Phillip and Zhu, Jun-Yan and Zhou, Tinghui and Efros, Alexei A},
  booktitle={Proceedings of the IEEE conference on computer vision and pattern recognition},
  pages={1125--1134},
  year={2017}
}

@inproceedings{MI2GAN,
  title={MI 2 GAN: generative adversarial network for medical image domain adaptation using mutual information constraint},
  author={Xie, Xinpeng and Chen, Jiawei and Li, Yuexiang and Shen, Linlin and Ma, Kai and Zheng, Yefeng},
  booktitle={International Conference on Medical Image Computing and Computer-Assisted Intervention},
  pages={516--525},
  year={2020},
  organization={Springer}
}

@article{jhu,
  title={Unsupervised reverse domain adaptation for synthetic medical images via adversarial training},
  author={Mahmood, Faisal and Chen, Richard and Durr, Nicholas J},
  journal={IEEE transactions on medical imaging},
  volume={37},
  number={12},
  pages={2572--2581},
  year={2018},
  publisher={IEEE}
}

@inproceedings{cltsGAN,
  title={CLTS-GAN: color-lighting-texture-specular reflection augmentation for colonoscopy},
  author={Mathew, Shawn and Nadeem, Saad and Kaufman, Arie},
  booktitle={International Conference on Medical Image Computing and Computer-Assisted Intervention},
  pages={519--529},
  year={2022},
  organization={Springer}
}

@article{c3vd,
  title={Colonoscopy 3D video dataset with paired depth from 2D-3D registration},
  author={Bobrow, Taylor L and Golhar, Mayank and Vijayan, Rohan and Akshintala, Venkata S and Garcia, Juan R and Durr, Nicholas J},
  journal={Medical image analysis},
  volume={90},
  pages={102956},
  year={2023},
  publisher={Elsevier}
}

@inproceedings{instructpix2pix,
  title={Instructpix2pix: Learning to follow image editing instructions},
  author={Brooks, Tim and Holynski, Aleksander and Efros, Alexei A},
  booktitle={Proceedings of the IEEE/CVF Conference on Computer Vision and Pattern Recognition},
  pages={18392--18402},
  year={2023}
}

@inproceedings{depthanything,
  title={Depth anything: Unleashing the power of large-scale unlabeled data},
  author={Yang, Lihe and Kang, Bingyi and Huang, Zilong and Xu, Xiaogang and Feng, Jiashi and Zhao, Hengshuang},
  booktitle={Proceedings of the IEEE/CVF Conference on Computer Vision and Pattern Recognition},
  pages={10371--10381},
  year={2024}
}

@article{shen2019context,
  title={Context-aware depth and pose estimation for bronchoscopic navigation},
  author={Shen, Mali and Gu, Yun and Liu, Ning and Yang, Guang-Zhong},
  journal={IEEE Robotics and Automation Letters},
  volume={4},
  number={2},
  pages={732--739},
  year={2019},
  publisher={IEEE}
}

@article{gu2022vision,
  title={Vision--kinematics interaction for robotic-assisted bronchoscopy navigation},
  author={Gu, Yun and Gu, Chuanjia and Yang, Jie and Sun, Jiayuan and Yang, Guang-Zhong},
  journal={IEEE Transactions on Medical Imaging},
  volume={41},
  number={12},
  pages={3600--3610},
  year={2022},
  publisher={IEEE}
}

@article{luo2023monocular,
  title={Monocular endoscope 6-DoF tracking with constrained evolutionary stochastic filtering},
  author={Luo, Xiongbiao and Xie, Lixin and Zeng, Hui-Qing and Wang, Xiaoying and Li, Shiyue},
  journal={Medical Image Analysis},
  volume={89},
  pages={102928},
  year={2023},
  publisher={Elsevier}
}

@inproceedings{luo2020new,
  title={A new electromagnetic-video endoscope tracking method via anatomical constraints and historically observed differential evolution},
  author={Luo, Xiongbiao},
  booktitle={Medical Image Computing and Computer Assisted Intervention--MICCAI 2020: 23rd International Conference, Lima, Peru, October 4--8, 2020, Proceedings, Part III 23},
  pages={96--104},
  year={2020},
  organization={Springer}
}

@article{zhu2024bronchoscopic,
  title={A bronchoscopic navigation method based on neural radiation fields},
  author={Zhu, Lifeng and Zheng, Jianwei and Wang, Cheng and Jiang, Junhong and Song, Aiguo},
  journal={International Journal of Computer Assisted Radiology and Surgery},
  volume={19},
  number={10},
  pages={2011--2021},
  year={2024},
  publisher={Springer}
}

@inproceedings{ace,
  title={Accelerated coordinate encoding: Learning to relocalize in minutes using rgb and poses},
  author={Brachmann, Eric and Cavallari, Tommaso and Prisacariu, Victor Adrian},
  booktitle={Proceedings of the IEEE/CVF Conference on Computer Vision and Pattern Recognition},
  pages={5044--5053},
  year={2023}
}

@inproceedings{zhu2023r2former,
  title={R2former: Unified retrieval and reranking transformer for place recognition},
  author={Zhu, Sijie and Yang, Linjie and Chen, Chen and Shah, Mubarak and Shen, Xiaohui and Wang, Heng},
  booktitle={Proceedings of the IEEE/CVF Conference on Computer Vision and Pattern Recognition},
  pages={19370--19380},
  year={2023}
}

@article{dsac,
  title={Visual camera re-localization from RGB and RGB-D images using DSAC},
  author={Brachmann, Eric and Rother, Carsten},
  journal={IEEE transactions on pattern analysis and machine intelligence},
  volume={44},
  number={9},
  pages={5847--5865},
  year={2021},
  publisher={IEEE}
}

@inproceedings{islam2024diffusemix,
  title={DiffuseMix: Label-Preserving Data Augmentation with Diffusion Models},
  author={Islam, Khawar and Zaheer, Muhammad Zaigham and Mahmood, Arif and Nandakumar, Karthik},
  booktitle={Proceedings of the IEEE/CVF Conference on Computer Vision and Pattern Recognition},
  pages={27621--27630},
  year={2024}
}

@inproceedings{pda,
  title={Camera pose matters: Improving depth prediction by mitigating pose distribution bias},
  author={Zhao, Yunhan and Kong, Shu and Fowlkes, Charless},
  booktitle={Proceedings of the IEEE/CVF Conference on Computer Vision and Pattern Recognition},
  pages={15759--15768},
  year={2021}
}

@inproceedings{monogs,
  title={Gaussian splatting slam},
  author={Matsuki, Hidenobu and Murai, Riku and Kelly, Paul HJ and Davison, Andrew J},
  booktitle={Proceedings of the IEEE/CVF Conference on Computer Vision and Pattern Recognition},
  pages={18039--18048},
  year={2024}
}

@inproceedings{zhang2023lite,
  title={Lite-mono: A lightweight cnn and transformer architecture for self-supervised monocular depth estimation},
  author={Zhang, Ning and Nex, Francesco and Vosselman, George and Kerle, Norman},
  booktitle={Proceedings of the IEEE/CVF Conference on Computer Vision and Pattern Recognition},
  pages={18537--18546},
  year={2023}
}

@article{gpt4-o,
  title={Gpt-4o system card},
  author={Hurst, Aaron and Lerer, Adam and Goucher, Adam P and Perelman, Adam and Ramesh, Aditya and Clark, Aidan and Ostrow, AJ and Welihinda, Akila and Hayes, Alan and Radford, Alec and others},
  journal={arXiv preprint arXiv:2410.21276},
  year={2024}
}

\end{document}